\documentclass[10pt,letterpaper,compsoc,conference]{iiswc22}
\usepackage{cite}
\usepackage{amsmath,amssymb,amsfonts}
\usepackage{algorithmic}
\usepackage{graphicx}
\usepackage[dvipsnames]{xcolor}
\usepackage[final]{microtype}
\usepackage[italic]{mathastext}
\usepackage{libertine}
\usepackage[T1]{fontenc}
\usepackage{textcomp}
\usepackage[varqu,varl]{zi4}
\usepackage[all]{nowidow}
\usepackage[auth-lg,affil-it]{authblk}
\usepackage[keeplastbox]{flushend}
\usepackage{formatting/packages}
\usepackage{formatting/names}
\usepackage{formatting/macros}
\usepackage{formatting/results}

\newcommand\blfootnote[1]{%
  \begingroup
  \renewcommand\thefootnote{}\footnote{#1}%
  \addtocounter{footnote}{-1}%
  \endgroup
}

\begin{document}

\title{An Evaluation of Edge TPU Accelerators for\\Convolutional Neural Networks}
\author[]{%
Kiran Seshadri$^1$\quad{}Berkin Akin$^*$\quad{}James Laudon$^{\ddagger{}}$\quad{}Ravi Narayanaswami$^{\dagger{}1}$\quad{}Amir Yazdanbakhsh$^{\ddagger{}}$\\
\small{Enfabrica\quad{}$^*$Google\quad{}$^\dagger$Cruise\quad{}$^{\ddagger{}}$Google Research, Brain Team}\\\vspace{0.3cm}
\scriptsize
{\texttt{\href{mailto:kiranks@utexas.edu}{kiranks@utexas.edu}, \href{mailto:bakin@google.com}{bakin@google.com}, \href{mailto:jlaudon@google.com}{jlaudon@google.com}},
\texttt{\href{mailto:ravi.narayanaswami@getcruise.com}{ravi.narayanaswami@getcruise.com},
\href{mailto:ayazdan@google.com}{ayazdan@google.com}
}}
}
\maketitle
\begin{abstract}
\label{sec:abstract}
Edge TPUs are a domain of accelerators for low-power, edge devices and are widely used in various Google products such as Coral and Pixel devices.
In this paper, we first discuss the major microarchitectural details of Edge TPUs.
Then, we extensively evaluate three classes of Edge TPUs, covering different computing ecosystems, across 423K unique convolutional neural networks.
Building upon this extensive study, we discuss critical and interpretable microarchitectural insights about the studied classes of Edge TPUs.
Mainly, we discuss how Edge TPU accelerators perform across convolutional neural networks with different structures.
Finally, we present a learned machine learning model with high accuracy to estimate the major performance metrics of accelerators.
These learned models enable significantly faster (in the order of milliseconds) evaluations of accelerators as an alternative to time-consuming cycle-accurate simulators and establish an exciting opportunity for rapid hardware/software co-design\blfootnote{$^1$~Work done when the authors were at Google.}.
\end{abstract}

\section{Introduction}
\label{sec:intro}
As a result of the diminishing returns in processor performance from the end of Moore’s Law, the last decade has seen a substantial surge in specialized hardware design.
This surge has provoked software-focused companies such as Google\cite{tpu,edgetpu}, Microsoft Brainwave~\cite{chung2018serving}, Amazon~\cite{amazon}, Apple~\cite{apple}, and Facebook~\cite{facebook} to invest heavily in designing specialized hardware to improve the efficiency of the compute underlying their core businesses.
In addition, the soaring demand for specialized hardware has also led to a fast-growing market for hardware startups.
Anticipating this trend, Google deployed Tensor Processing Units (TPUs) in 2015 to accelerate machine learning inference in data centers.
Two years later, in 2017 Google introduced TPUv2~\cite{tpu} to accelerate machine learning training.
Following the TPUs, Google debuted Edge TPU accelerators~\cite{edgetpu}, the focus of this paper, in 2018 for machine learning inference at the edge.
Edge TPUs primarily target delivering high performance acceleration within tight physical and power budgets.
Since their debut, Edge TPUs have been used in various Google products such as Coral~\cite{coral} and Pixel phones~\cite{pixel}.
The Edge TPU ecosystem is built with full parameterization across the computing stack which enables various design space exploration of architecture configurations.

Concretely, we outline the contributions of our paper as follows:
\begin{itempacked}
\item \niparagraph{Evaluating three classes of Edge TPUs across large numbers of convolutional neural networks.}
We evaluate three classes of Edge TPUs using nearly 423K convolutional neural networks~\cite{nasbench101} with diverse structures and various convolution operations. 
The studied Edge TPUs embody accelerators that are either already deployed in recent Google products\cite{coral,pixel} or in the pipeline to be used in future products.
\item \niparagraph{Outlining critical architectural insights about Edge TPUs.}
Analyzing the evaluation results, we outline critical insights about the Edge TPU architectures and how they perform across various convolutional neural networks.
Particularly, we outline how these classes of architectures work across convolutional neural models with different sizes and structures.
In addition, we explain how the accelerator tile size impact the performance of the Edge TPU accelerators.
Finally, we show the deltas in the accelerator performance after replacing an operation with another operation. 
\item \niparagraph{Developing efficient and robust learned models to estimate major performance metrics of Edge TPUs.}
We also discuss our proposed high-accuracy learned model for estimating various performance metrics of Edge TPUs.
We use graph neural networks to learn a latent representation of input graphs and estimate the desired performance metrics.
Our initial results show that the learned model estimates the critical performance metrics of the workloads with around 3$\%$ estimation error and significantly high (around 0.99) rank-based correlation metric with ground truth data.
This strong correlation signifies that the introduced learned performance is a strong candidate for replacing the expensive-to-evaluate cycle-accurate simulators in design space exploration and hardware/software co-optimization.
\end{itempacked}
\section{Edge TPU Microarchitecture}
\label{sec:microarch}
\begin{figure}[t]
    \centering
    \includegraphics[width=0.48\textwidth]{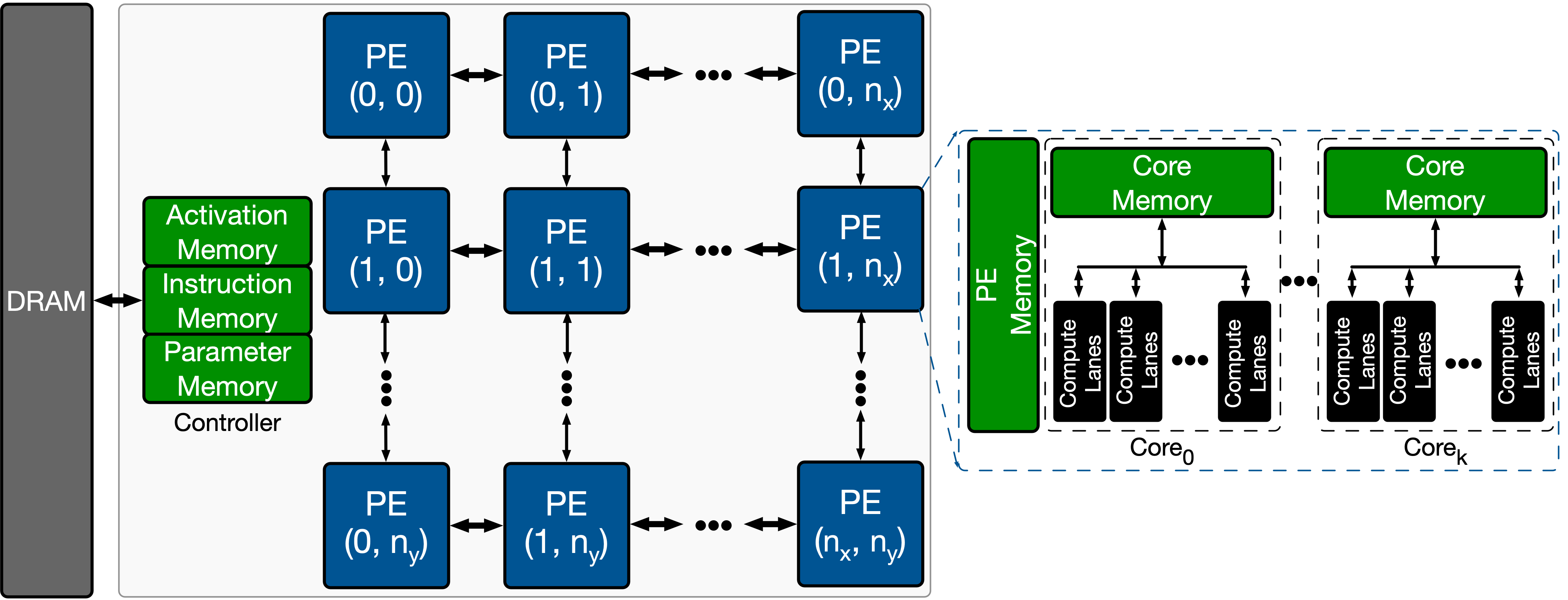}
    \caption{Overview of the template-based machine learning accelerator used for architecture exploration.}
    \label{fig:template_accel}
\end{figure}

Figure~\ref{fig:template_accel} shows the overall architecture of Edge TPU accelerators.
The Edge TPU accelerators leverage a template-based design with highly parameterizable microarchitectural components.
The parameterized design of Edge TPU accelerators enable exploring various architecture configurations for different target applications.
As shown in Figure~\ref{fig:template_accel}, the template accelerator is organized in a 2D array of processing elements (PEs).
Each PE performs a set of arithmetic computations in a single instruction multiple data (SIMD) manner.
An on-chip controller is used to transfer the data from off-chip memory and PEs.
The controller fetches activation and parameters into the on-chip staging buffers.
In addition, the controller reads in the low-level instructions (e.g. convolution, etc.) that will be executed on the PEs.

The main architectural components of each processing engine are a single or multiple core(s) each with multiple compute lanes for performing operations in SIMD manner.
Following a top-down approach, each PE has a memory shared across all the compute cores.
This memory, shown as PE memory in Figure~\ref{fig:template_accel}, is mainly used to store model activations, partial results, and outputs.
The cores within each PE feature a \textit{core memory} that is mainly used for storing model parameters.
Each core has multiple compute lanes where each lane has multi-way multiply-accumulate (MAC) units.
The core memories are heavily multi-banked to keep up with the compute throughput of the parallel compute lanes and their SIMD MAC units.
At each cycle, a set of activations are sent to the compute lanes.
Then, the computations between the activations and model parameters are performed within each lane using the multi-way MAC units.
Once the computations finish, the results are either stored back in the PE memory for further computation or are offloaded back into the DRAM.

\section{Edge TPU Software Ecosystem}
\label{sec:sw}
\begin{figure*}[t]
    \centering
    \includegraphics[width=0.98\textwidth]{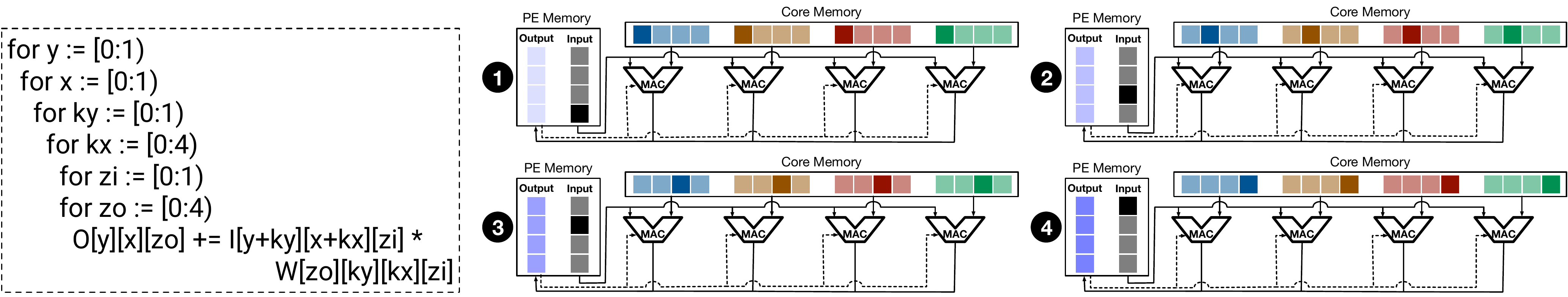}
    \caption{Convolutional layer loop nest (without batching) and its mapping on an Edge TPU with one PE and four SIMD lanes (\# MACs). Input and output activations are stored in PE memory. The weight parameters are stored in core memory.}
    \label{fig:mapping}
\end{figure*}

In this work we use TensorFlow Lite~\cite{tflite} based models as input to the Edge TPU software stack.
Note that the software ecosystem varies based on the particular Edge TPU platform but~\cite{edgeruntime} provides an overview of running TensorFlow Lite based inferences for the Coral Edge TPU platforms.
The Edge TPU runtime library is used to communicate with the accelerator from the TensorFlow Lite (TFLite) API~\cite{edgeruntime} where the
TFLite models are compiled ahead-of-time using a publicly available Edge TPU compiler~\cite{edgecompiler}.
The main goal of the compiler is to map various neural network operations supported on the Edge TPU hardware while extracting the highest level of parallelism.
Note that if the input models have unsupported or non-quantized operations, the compiler partitions the input graphs where the unsupported portion runs on a CPU instead of the Edge TPU. 

\niparagraph{Parameter caching.}
One critical optimization that the compiler performs is parameter caching~\cite{edgecompiler}.
As on-chip memory size is generally scarce on edge accelerators, efficiently managing this scratch-pad space is essential.
For continuous inference scenarios, reloading the entire neural network model for every inference results in significant time spent on the memory transfers.
If the parameters (i.e. weights) of the neural network are fully or partially cached in the on-chip memory, consecutive inferences on the new inputs can reuse the cached parameters.
This process significantly reduces external memory transfers which leads to higher performance and energy efficiency.

\niparagraph{Mapping of convolution operation into accelerator.}
Figure~\ref{fig:mapping} shows the loop nest of a convolutional layer (without batching) and its mapping onto an Edge TPU accelerator with one processing engine and four SIMD lanes.
As depicted in Figure~\ref{fig:mapping}, the PE memory is used for both input activations, partial sums, and outputs.
On the other hand, the core memory is exploited to store the weight parameters.
Depending on the size of PE memory, core memory, input activations, weight parameters, and outputs, the compiler~\cite{edgecompiler} may choose a different mapping and tiling for the data.

Figure~\ref{fig:mapping} depicts an example showing the computation steps for performing a convolution operation on an Edge TPU accelerator.
The squares with darker color show the data elements that are active at each particular computation step, whereas the squares with light color show the data elements that are inactive at a computation step.
In this example, the input activation (in PE memory) has four elements.
There are four weight parameters, each mapped onto a different core.
The weight parameters also have four elements.
Finally, the convolutional window size is (K$_x$ = 1, K$_y$ = 4, Z$_i$ = 1), and to compute one output element it requires all four elements of the input activation to be multiply-and-accumulated with all four elements of the weight parameters.
Hence, in the loop nest representation, the $x$ and $y$ variables are set to one.
At the first iteration (\circled{1}), the first element of the input activation and the first element of each of the (Zo = 4) kernels are multiplied together.
The partial sum value is stored in the output array.
Then (\circled{2}), the second element of the input activation and the second element of each kernel are multiplied together. The multiplication result is accumulated with the previous partial sum and stored back into the output array.
The computation continues until all the elements of the input activation are processed (\circled{4}).
\section{Learned Performance Model}
\label{sec:gnn}
In this paper, we use graph neural networks~\cite{kipf2016semi} to learn a generalized model for the performance of Edge TPUs across different convolutional neural networks.
This work explores generalization across different neural network architectures and shows the feasibility of employing unique machine learning models for performance prediction across different accelerators.
We use an in-house detailed cycle-accurate performance model to collect training data for the learned model (See Section~\ref{sec:methodology} for details).
Using graph neural networks, we learn a latent representation for each convolutional neural network.
This latent representation is later used to estimate various performance metrics (e.g. latency and energy).

A convolutional neural network can be presented with a graph $g = (V, E)$, where $V$ is a set of nodes that represent the valid operations (e.g. 3$\times$3 convolution, 1$\times$1 convolution, 3$\times$3 max-pooling, input, and output), $E$ is a set of edges that represent the connectivity between the operations, and $G$ that is a vector serving as a graph global feature.
A graph neural network takes as input the feature description of the nodes, an adjacency matrix, and a global feature of the graph and performs multiple iterations of message-passing~\cite{kipf2016semi} to learn a node-level representation.
The learned node representations are aggregated into a graph-level representation using arithmetic operators such as summation or averaging~\cite{hamilton2017inductive,graphnet}.
This graph-level representation is used as the performance model predictor.
In the next paragraph, we elaborate the structure of our graph-based performance predictor.

\subsection{Learned Performance Model Structure}
\begin{figure}[t]
    \centering
    \includegraphics[width=0.45\textwidth]{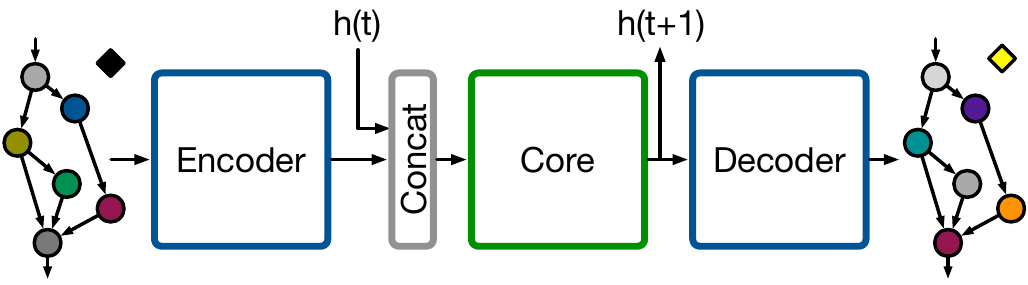}
    \caption{High-level overview of the graph-based learned performance model. The model consists of three main components: encoder, core, and decoder. The encoder and decoder independently perform computations on edge, node, and global features, whereas the core component that performs multiple rounds of message-passing to map the input graph structural dependencies into a latent representation. The disconnected tilted square represents the global feature (single float scalar) of the graph. After update, the global feature is used as the predicted performance metric (e.g. latency, energy, etc.).}
    \label{fig:perfmodel}
\end{figure}
Recent work has explored employing various machine learning or analytical models for performance estimation~\cite{critical_path:pmbs:2019,iaca:web:2019,llvm-mca:web:2019,throughput_intel:pmbs:2018,zsim:archnews:2013,marss:dac:2011,thread:hpca:2009,throughput_sc:tc:2008,chronos:scp:2007,perf_parallel:tocs:2004,pred_parallel:pc:2004,wcet:es:2001,pred_parallel:ipds:1998,static_perf:sc:1993,pred_static:rts:1993,bound:tools:1992,tpucost:arxiv:2020,halide:tog:2019,ithemal:icml:2019,quant_game:tecs:2012,gametime:tools:2011,regression:nips:2010,compiler_perf:cf:2007,granite:iiswc:2022,yazdanbakhsh2021apollo,hegde2021mind,zhou2022towards,kumar2021data} of different applications and/or hardware accelerators.
In this work, to implement the learned performance model, we use DeepMind's Graph Nets~\cite{graphnet} and Sonnet~\cite{sonnet} libraries.
We use a graph-based model because (a) application mapping is more straightforward and (b) graph neural networks generally better capture the dependencies between nodes~\cite{tpucost:arxiv:2020,shi2019learning}.
Figure~\ref{fig:perfmodel} depicts the overall structure of the graph-based learned performance model.
The model consists of three main components, namely \textit{encoder}, \textit{core}, and \textit{decoder}, each serving as a neural network model.
The encoder (decoder) network independently encodes (decodes) the edge, node, and global attributes. 
Note that, neither encoder nor decoder compute the structural relations between the graph components.
The core component of the model performs multiple rounds of message-passing steps.
After the core component computations complete, the structural relations between the graph components are learned and mapped into a node-level and/or graph-level representation.
\begin{figure}[t]
    \centering
    \subfloat[Sample NASBench-101 cell.]{
    \includegraphics[width=0.23\textwidth]{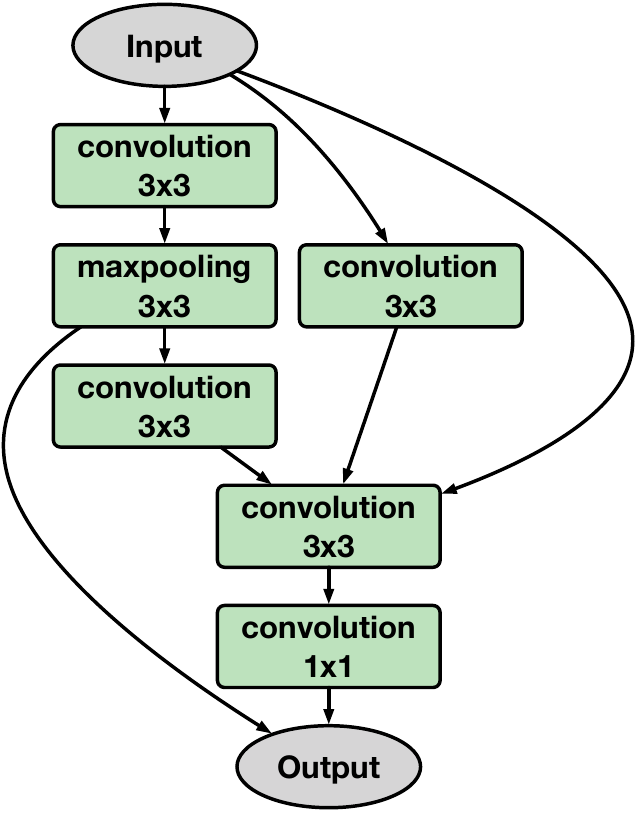}
    \label{fig:orig_g}}
    \subfloat[Node, edge, and global features.]{
    \includegraphics[width=0.23\textwidth]{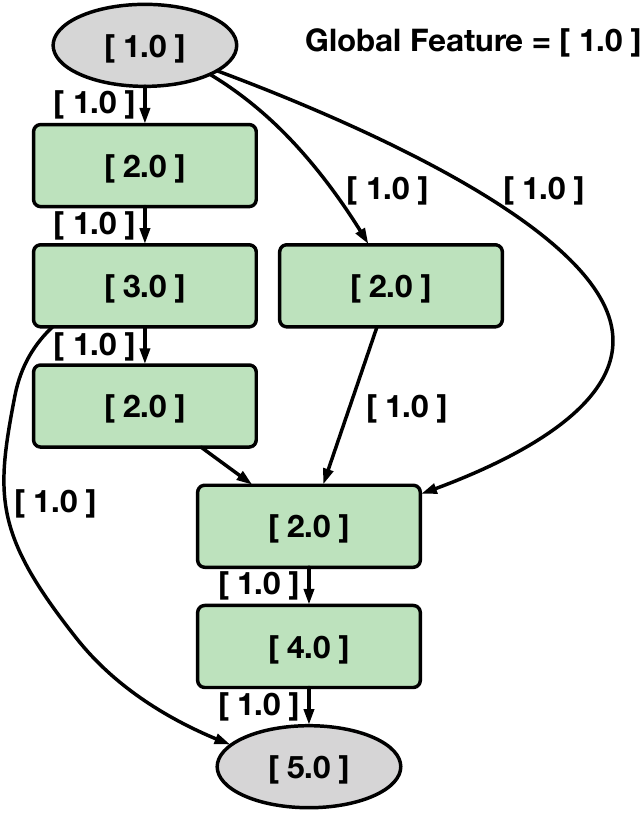}
    \label{fig:trans_g}}
    \caption{(a) A sampled cell from NASBench-101~\cite{nasbench101} dataset with six operation nodes and nine edges. (b) The representation of node, edge, and global features. In this mapping, input and output nodes are mapped to feature vectors [1.0] and [5.0], respectively. The operation 3$\times$3 convolution, 3$\times$3 max-pooling, and 1$\times$1 convolution are mapped to feature vectors [2.0], [3.0], and [4.0], respectively. Both edge and global features are set to feature vector [1.0].} 
    \label{fig:graph_repr}
\end{figure}

\niparagraph{Input graph representation.}
Figure~\ref{fig:graph_repr} shows a randomly selected cell from one of the convolutional neural network models in NASBench-101~\cite{nasbench101} and its corresponding node, edge, and global feature vectors.
We employ a simple float-encoding for the valid operations in the graph.
We map input, 3$\times$3 convolution, 3$\times$3 max-pooling, 1$\times$1 convolution, and output nodes to feature vector [1.0] to [5.0], respectively.
In our current mapping, we do not differentiate between the edges and set all the edge features to [1.0].
Finally, for the graph-level representation, we use global feature vector [1.0].

\niparagraph{Encoder/decoder components.}
The encoder/decoder components independently apply neural network models to edge, node, and global attributes.
In our learned performance model, we use two-layered feed-forward neural networks each with 16 neurons, followed by a layer normalization~\cite{ba2016layer} for edge, node, and global features.

\niparagraph{Core component.}
The core component, as its name indicates, is the core computation part of the learned performance model.
To better understand the architecture of the core component, we summarize the Graph Net library~\cite{graphnet} block-structure concept that is used as building block for implementing flexible graph neural networks.
Graph Net~\cite{graphnet} consists of three main neural model blocks, namely edge, node, and global.
The edge, node, and global neural model blocks use edge, node, and global attributes as their output.
That is, after applying each neural model block, the corresponding output attribute gets updated.
Using this methodology, various graph neural network architectures can be implemented simply by gluing these neural model blocks together.
Generally, the computation steps in a full graph neural network block (See Algorithm 1~\cite{graphnet}) is as follows: 

\begin{itempacked}
\item \niparagraph{Edge update.} Apply the edge neural model block on each edge and update its attribute based on the previous edge features, the features of the adjacent nodes, and the global feature of the graph;
\item \niparagraph{Node update.} Apply the node neural model block on each node and update its attribute based on the previous node features, the features of incoming edges, and the global feature of the graph;
\item \niparagraph{Global update.} Apply the global neural model block on the global feature and update its attribute based on the previous global feature, globally aggregated features of the edges, and globally aggregated features of the nodes.
\end{itempacked}
The inputs to each neural model block and the aggregation type can be tailored to the demands of the target task. 
In our implementation, we use the default configurations for edge, node, and global blocks.
In addition, we use the default summation operation as the aggregator for edge, node, and global neural model blocks.
We use the final updated global attribute (a single float scalar) as the predicted performance metric (e.g. latency, energy, area, etc.).
\section{Methodology}
\label{sec:methodology}
\begin{table}
\footnotesize
\centering
\aboverulesep=0ex 
\belowrulesep=0ex 
\caption{The distribution of NASBench-101~\cite{nasbench101} models across different intervals of trainable parameters.}
\label{table:dist_model}
\begin{tabular}{l|l}
\toprule
\textbf{Trainable Parameters Intervals} & \textbf{$\#$ of Models}\\\toprule
$[$227,274 --- 5,202,474$)$&210,673\\\midrule
$[$5,202,474 --- 10,177,674$)$& 102,488\\\midrule
$[$10,177,674 --- 15,152,874$)$ & 44,272\\\midrule
$[$15,152,874 --- 20,128,074$)$ & 3,513\\\midrule
$[$20,128,074 --- 25,103,274$)$& 38,003\\\midrule
$[$25,103,274 --- 30,078,474$)$& 4,413\\\midrule
$[$30,078,474 --- 35,053,674$)$& 15,041\\\midrule
$[$35,053,674 --- 40,028,874$)$& 3,533\\\midrule
$[$40,028,874 --- 45,004,074$)$& 1,209\\\midrule
$[$45,004,074 --- 49,979,274$)$& 479\\\bottomrule
\end{tabular}
\end{table}
\niparagraph{Workloads.}
We use NASBench-101~\cite{nasbench101} dataset that include nearly 423K unique convolutional neural network architectures with diverse structures and various number of convolutional operations.
NASBench dataset are widely used in AutoML evaluation efforts~\cite{automl:data,automl:bananas,automl:metaarch,automl:efffwd,automl:prdictor,automl:graph}.
The valid operations in these neural network architectures are 3$\times$3 convolution, 1$\times$1 convolution, and 3$\times$3 max-pooling.
The neural networks in the NASBench dataset consist of three stacks followed by a downsampling layer.
Each stack has a repeated structure of \textit{cell}s.
The space of cell architecture includes all possible directed acyclic graph combinations of valid operations while complying with input/output dimensional constraints. 
To restrict the search space, the maximum number of vertices and edges within each cell are set to seven and nine, respectively. 
The NASBench dataset also includes evaluated performance (e.g. training and validation accuracy) of the neural network architectures on CIFAR-10 image dataset~\cite{cifar10} at different epochs (4, 12, 36, and 108) totalling $\sim$5M trained models.
This enables studying the trade-offs between the performance of Edge TPUs and training/validation accuracy of neural network models.
Table~\ref{table:dist_model} illustrates the distribution of models across different intervals of trainable parameters, which covers a diverse set of model sizes with different characteristics (e.g. compute- and memory-intensive).
\begin{figure*}[!t]
    \centering
    \subfloat[V1]{
    \includegraphics[width=0.33\textwidth]{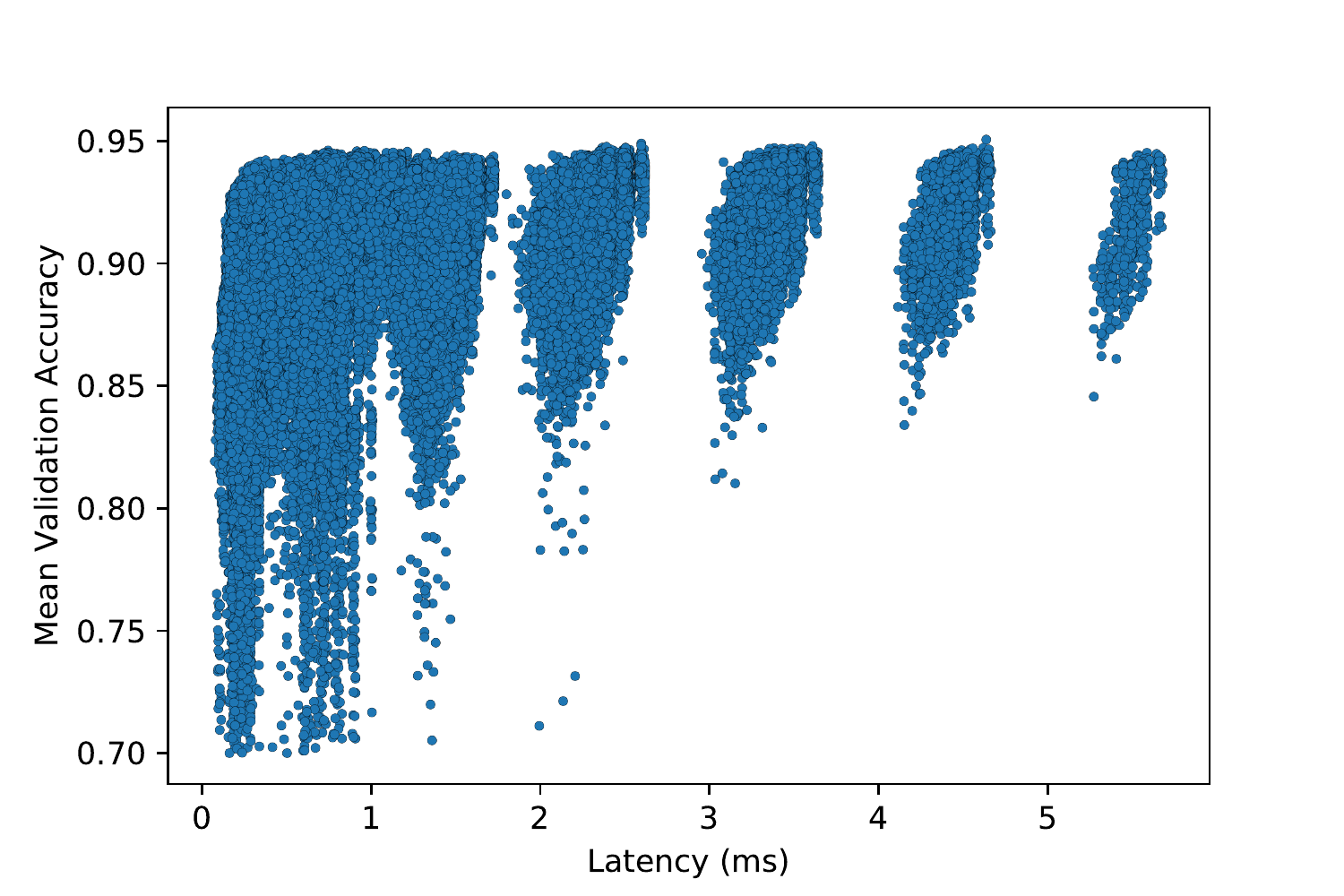}
    \label{fig:dc_lat_acc}}
    \subfloat[V2]{
    \includegraphics[width=0.33\textwidth]{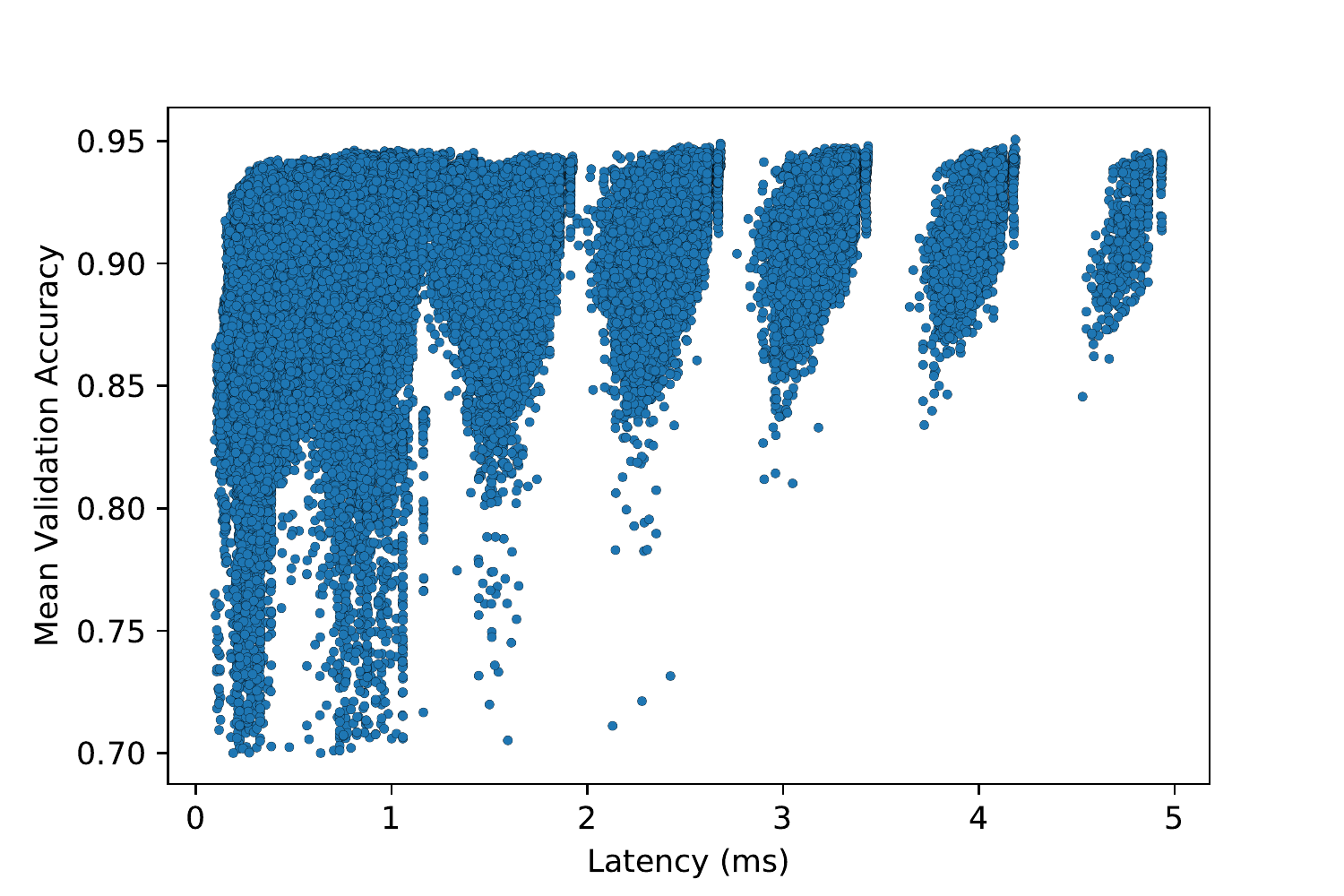}
    \label{fig:msoc1_lat_acc}}
    \subfloat[V3]{
    \includegraphics[width=0.33\textwidth]{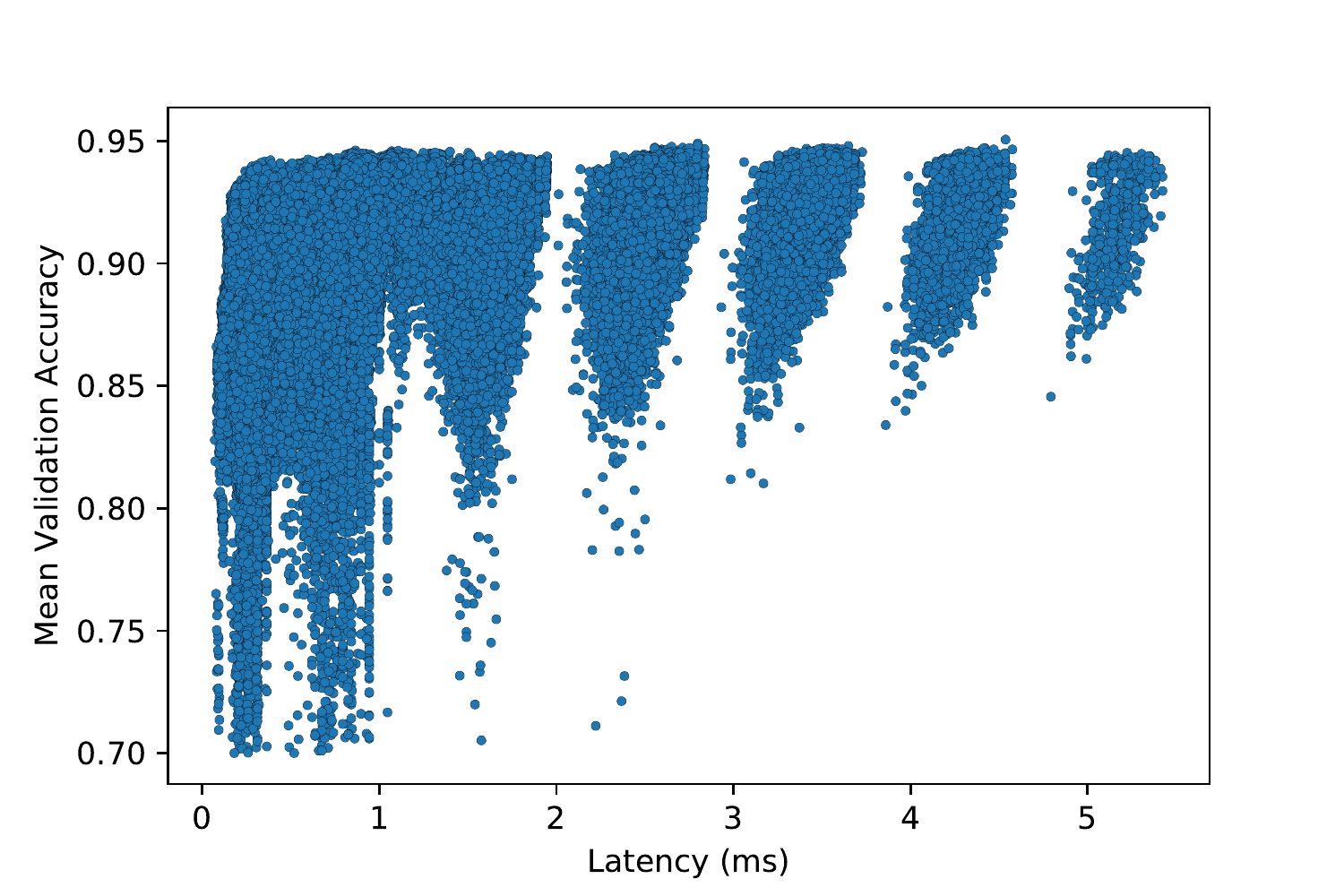}
    \label{fig:msoc2_lat_acc}}
    \caption{The scatter plot of mean validation accuracy at epoch 108 vs. latency for three Edge TPU classes, (a) configuration V1, (b) configuration V2, and (c) configuration V3.} 
    \label{fig:lat_acc}
\end{figure*}

\niparagraph{Accelerator configurations.}
Table~\ref{tab:archs} depicts several key microarchitectural features of the studied classes of Edge TPU accelerators.
The system clock frequency of V1, V2, and V3 accelerators are 800~MHz, 1066~MHz, and 1066~MHz, respectively.
Comparing the V2 and V3 accelerators, V3 has larger PE memory (2~MB vs. 384~KB), more cores per PE (8 vs. 1), while using fewer Y-PEs (1 vs. 4).
\begin{table}[t]
  \setlength{\tabcolsep}{2pt}
  \begin{center}
    \caption{The detailed microarchitecture parameters of three studied configurations of Edge TPU accelerators, covering various computing ecosystems. The total core memory capacity per accelerator is equal to core memory $\times$ number of PEs $\times$ number of cores. Each class covers different domain: V1~$\mapsto$~high peak TOPs, V2~$\mapsto$~low peak TOPs with small on-chip memory, and V3~$\mapsto$~low peak TOPs with large on-chip memory.}
    \label{tab:archs}
    \begin{tabular}{l|P{1.4cm}|P{1.9cm}|P{1.9cm}|}
    \cline{2-4}
    & V1 & V2 & V3\\\hline
    \multicolumn{1}{|l|}{\textbf{Clock Frequency (MHz)}}&800&1066&1066\\\hline
    \multicolumn{1}{|l|}{\textbf{\# of (X, Y)-PEs}}&(4, 4)&(4, 4)&(4, 1)\\\hline
    \multicolumn{1}{|l|}{\textbf{PE Memory}}&2~MB&384~KB&2~MB\\\hline
    \multicolumn{1}{|l|}{\textbf{\# of Cores per PE}}&4&1&8\\\hline
    \multicolumn{1}{|l|}{\textbf{Core Memory}}&32~KB&32~KB&8~KB\\\hline
    \multicolumn{1}{|l|}{\textbf{\# of Compute Lanes}}&64&64&32\\\hline
    \multicolumn{1}{|l|}{\textbf{Instruction Memory}}&16384&16384&16384\\\hline
    \multicolumn{1}{|l|}{\textbf{Parameter Memory}}&16384&8192&8192\\\hline
    \multicolumn{1}{|l|}{\textbf{Activation Memory}}&1024&1024&1024\\\hline
    \multicolumn{1}{|l|}{\textbf{I/O Bandwidth (GB/s)}}&17&32&32\\\hline
    \multicolumn{1}{|l|}{\textbf{Peak TOPS}}&26.2&8.73&8.73\\\hline
    \end{tabular}
  \end{center}
\end{table}

\niparagraph{Microarchitectural simulations.}
To provide a uniform simulation infrastructure across all the the accelerator configurations, we use an in-house fully-parameterized cycle-accurate performance model.
The simulator measures the latency and energy of running the workloads for different accelerator configurations.
To provide highest performance, in all the simulations, we enable parameter caching~\cite{edgecompiler}.

\niparagraph{Learned performance model training.}
We train our graph network model with the Adam optimizer~\cite{adam:arxiv:2014} using default parameters and learning rate of 1e-3.
We randomly split the dataset (423K data points) into 60$\%$ training, 20$\%$ validation, and 20$\%$ testing.
We use DeepMind Sonnet~\cite{sonnet} and Graph Nets~\cite{graphnet} library to implement the neural models.
For edge, node, and global neural model blocks, we use two-layer feed-forward models with 16 neurons at each layer, followed by a normalization layer.
Because of its simple architecture, the latency of the learned model to generate a prediction is on the order of milliseconds.
We use the default zero initializer for the bias.
For the weights, we use truncated random normal values with a standard deviation proportional to the number of input features~\cite{batchnorm}.
For the normalization layer, we create variables to hold the scale and offset of the normalization.
For the training loss, we compute the mean square of the prediction error from every iteration of the message passing in order to enable the model to converge faster across every iteration of message passing.
In this work, we train separate GNN models per class of Edge TPU.
Since NASBench-101 is constructed by repeated same cell architecture, for each NASBench-101 model we use the cell architecture as the input to the learned model\footnote{The source code of our implementations can be found under open-source license at \scriptsize{\url{https://github.com/google-research/google-research/tree/master/l2da}}.}.
\begin{table}[t]
  \setlength{\tabcolsep}{3pt}
  \begin{center}
    \caption{The summary of latency and energy measurements of the convolutional neural networks~\cite{nasbench101} with at least 70$\%$ mean validation accuracy (417,454 out of 423,624 NASBench models---$\approx$98.5$\%$ of total data points) across three classes of Edge TPUs (See Table~\ref{tab:archs}). The values in parentheses shows the corresponding mean validation accuracy.}
    \label{tab:summary:results}
    \begin{tabular}{P{1.cm}|P{1.5cm}|P{1.9cm}|P{1.9cm}|}
    \cline{2-4}
    &\textbf{V1}&\textbf{V2}&\textbf{V3}\\\hline
    \multicolumn{1}{|l|}{\textbf{Min. Latency~(\si{\milli\second})}}&\shortstack{0.079111 \\(81.94$\%$)}&\shortstack{0.074647\\(82.81$\%$)}&\shortstack{0.074647\\(82.81$\%$)}\\\hline
    \multicolumn{1}{|l|}{\textbf{Max. Latency~(\si{\milli\second})}}&\shortstack{5.676561\\(93.78$\%$)}&\shortstack{5.653848\\(94.33$\%$)}&\shortstack{5.666214\\(93.55$\%$)}\\\hline
    \multicolumn{1}{|l|}{\textbf{Avg. Latency~(\si{\milli\second})}}&\shortstack{0.9631\\(N/A)}&\shortstack{1.03485\\(N/A)}&\shortstack{1.0655\\(N/A)}\\\hhline{|=|=|=|=|}
    \multicolumn{1}{|l|}{\textbf{Min. Energy~(\si{\milli\joule})}}&\shortstack{0.198351\\(81.94$\%$)}&\shortstack{0.170954\\(81.94$\%$)}&N/A\\\hline
    \multicolumn{1}{|l|}{\textbf{Max. Energy~(\si{\milli\joule})}}&\shortstack{23.807941\\(93.66$\%$)}&\shortstack{23.462845\\(92.97$\%$)}&N/A\\\hline
    \multicolumn{1}{|l|}{\textbf{Avg. Energy~(\si{\milli\joule})}}&\shortstack{4.252673\\(N/A)}&\shortstack{3.9127185\\(N/A)}&N/A\\\hline
    \end{tabular}
  \end{center}
\end{table}
\begin{table}[t]
  \setlength{\tabcolsep}{2pt}
  \begin{center}
    \caption{The latency and energy measurements of the neural network with maximum accuracy (95.055$\%$) after 108 epochs of training across three classes of Edge TPUs (See Table~\ref{tab:archs}). At the time of submission, the energy model for V3 was not available.}
    \label{tab:summary:best_acc}
    \begin{tabular}{P{1.cm}|P{1.5cm}|P{1.9cm}|P{1.9cm}|}
    \cline{2-4}
    &\textbf{V1}&\textbf{V2}&\textbf{V3}\\\hline
    \multicolumn{1}{|l|}{\textbf{Latency~(\si{\milli\second})}}&4.633768&4.185697&4.535305\\\hhline{|=|=|=|=|}
    \multicolumn{1}{|l|}{\textbf{Energy~(\si{\milli\joule})}}&19.894033&19.745373&N/A\\\hline
    \end{tabular}
  \end{center}
\end{table}
\section{Evaluation}
\label{sec:eval}
\niparagraph{Inference latency and energy measurements.}
We perform inference latency measurements for three different configurations of the Edge TPU accelerator across all NASBench-101~\cite{nasbench101} models (423K models) yielding a total number of latency measurements of approximately 1.5 Million (3$\times$423K).
The inference latency numbers measure the total inference time including off-chip accesses.
Similarly, we measure total energy for the \bench{V1} and \bench{V2} configurations across all the NASBench models (423K models) yielding a total number of energy measurements of approximately 900K (2$\times$423K). At the time of submission, the energy model for \bench{V3} was not available.

Table~\ref{tab:summary:results} shows the summary of the latency and energy results for three studied Edge TPU configurations and across the NASBench models with at least 70$\%$ mean validation accuracy after 108 training epochs.
The number of data points after this filtering is 417,454; more than 98.5$\%$ of the total number of data points (423K).
Compared to other accelerator configurations, \bench{V2} performs better in terms of delivering the highest accuracy (94.33$\%$) with lower latency (5.65 ms).
The high performance of this Edge TPU configuration is mainly attributed to its larger core memory (32~KB) and higher I/O bandwidth (32~GB/s).
The results also depict an interesting trend for the average energy consumption between \bench{V1} and \bench{V2}. 
Compared to class \bench{V1}, \bench{V2} has less PE memory and more I/O bandwidth (Table~\ref{tab:archs}), which may indicate more off-chip memory accesses.
We attribute this trend to \bench{V1}'s lower clock frequency and its lower average number of cycles to execute the models (1,584,211.2 in \bench{V1} vs. 2,080,014.9 in \bench{V2}).

Table~\ref{tab:summary:best_acc} shows the latency and energy consumption of the best model in terms of mean validation accuracy.
The mean validation accuracy (defined in~\cite{nasbench101}) is calculated across the three repeats of training after 108 training epochs~\cite{nasbench101}.
The results corroborate our initial observation that \bench{V2} is more efficient compared to other configurations.

Figure~\ref{fig:lat_acc} shows the comparison between mean validation accuracy and accelerator latency across all the convolutional models with at least 70$\%$ mean validation accuracy.
As we can see, the data are clustered into different buckets.
The number of 3$\times$3 convolution operations in each NASBench cell is a key determinant of latency buckets.
For example, an increase of one 3$\times$3 convolution operation yields a jump from one latency bucket to the next one.
The first three buckets (latency of $<$ 2.0~ms, 2.0$-$3.0~ms, and 3.0$-$4.0~ms, respectively) contain NASBench cells with respective averages of 1.48, 2.0, and 3.0 3$\times$3 convolution operations.

\begin{figure}[!t]
    \centering
    \includegraphics[width=0.48\textwidth]{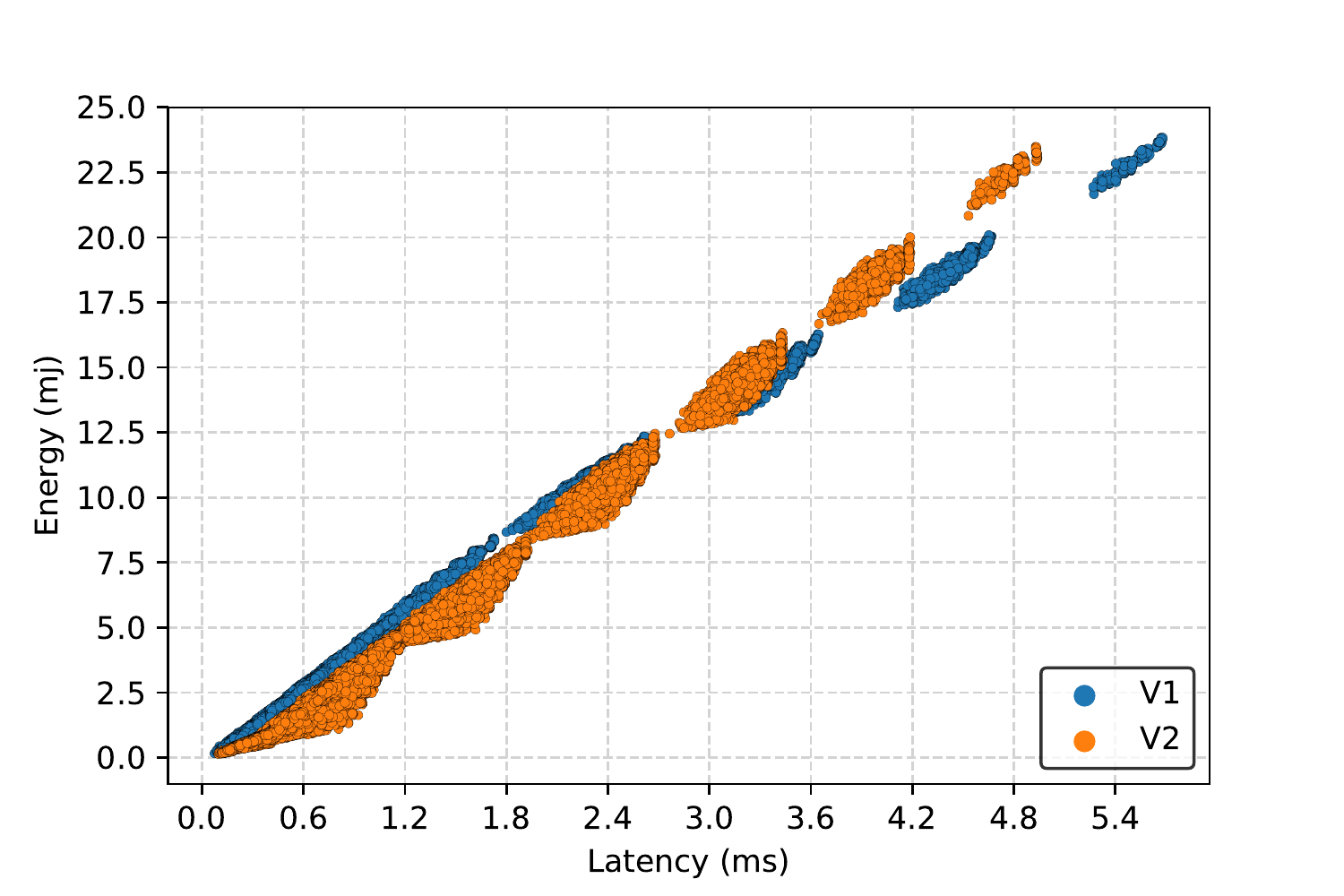}
    \caption{Scatter plot of the measured inference latency vs. the inference energy for the NASBench models with at least 70$\%$ mean validation accuracy on V1 and V2 configurations. At the time of submission, the energy model for V3 was not available.}
    \label{fig:latency_vs_energy}
\end{figure}

Figure~\ref{fig:latency_vs_energy} shows the relationship between the measured inference latency and inference energy for the NASBench models with at least 70$\%$ mean validation accuracy on the \bench{V1} and \bench{V2} configurations.
The relationship between the latency and energy is linear.
The first observation from Figure~\ref{fig:latency_vs_energy} is that for models with low latency ($<$ 3.0 ms), \bench{V2} yields lower energy compared to \bench{V1}.
However, as latency increases (models with larger number of parameters), \bench{V1} consumes less energy.
This trend can be attributed to the large PE memory size in \bench{V1} that enables running large models more efficiently. However, since \bench{V2} has smaller PE memory (only 384~KB), it may trigger multiple buffer flush/refills for these models.
\begin{table*}[ht]
  \setlength{\tabcolsep}{2pt}
  \begin{center}
    \caption{The average measured inference latency and energy of neural models on every configuration. Latency(X) $\leq$ contains all the neural models whose measured inference latency is the least on the Edge TPU configuration X. At the time of submission, the energy model for V3 was not available.}.
    \label{tab:summary:bucket}
    \begin{tabular}{P{1.8cm}|P{2cm}|P{1.9cm}|P{1.9cm}|P{1.9cm}|}
    \cline{2-5}
    &\multirow{2}{*}{\textbf{\# of Models}}&\multicolumn{3}{|c|}{\textbf{Average Latency (\si{\milli\second}), Average Energy (\si{\milli\joule})}}\\\cline{3-5}
    &&\textbf{V1}&\textbf{V2}&\textbf{V3}\\\hline
    \multicolumn{1}{|l|}{\textbf{Latency(V1) $\leq$}}&392725&0.80, 3.58&0.90, 3.58&0.92, N/A\\\hline
    \multicolumn{1}{|l|}{\textbf{Latency(V2) $\leq$}}&24325&3.73, 6.96&3.39, 15.67&3.61, N/A\\\hline
    \multicolumn{1}{|l|}{\textbf{Latency(V3) $\leq$}}&6570&2.59, 0.85&0.31, 0.64&0.25, N/A\\\hline
    \end{tabular}
  \end{center}
\end{table*}

\niparagraph{Performance comparison of the accelerators.}
To compare the performance of the neural models across different configurations, we split the NASBench models into three buckets, one per accelerator configuration. 
Each bucket contains all the models whose measured latency is the least for that particular configuration, as shown in Table~\ref{tab:summary:bucket}.
For example, first row of the table shows all the models whose measured inference latency is the least on \bench{V1} compared to other accelerator configurations.
First column of the table shows the number of models out of total NASBench models (423K) that reside in the corresponding bucket.
Overall, \bench{V1} performs better (larger number of models) compared to other configurations in terms of latency.
The next three columns show average latency and average energy of across the models that belong to the bucket for all three configurations.
On average, for the models with higher latencies (second row), \bench{V2} performs better.
Last bucket, in which \bench{V3} performs better compared to other accelerator configurations, mainly consists of models with a larger number of 1$\times$1 convolution.
On average, for the models in this bucket \bench{V3} yields 10.4$\times$ and 1.24$\times$ speedup compared to \bench{V1} and \bench{V2}, respectively.
The performance disparity between Edge TPU classes can be attributed to multiple application and/or accelerator characteristics.
Table~\ref{table:bucketconv} highlights some of the differences between first and last bucket with respect to the model characteristics.
The first bucket contains models with more parameters (higher memory-boundedness), due to more 3$\times$3 convolution operations.
\bench{V1} performs better because of its higher peak TOPs and larger on-chip memory. 
On the other hand, the last bucket contains models with fewer parameters. 
The differences between the frequency of \bench{V1} (800\,MHz), \bench{V2} (1066\,MHz), and \bench{V3} (1066\,MHz) also contribute to the overall performance of each bucket.
\begin{table}
\scriptsize
\centering
\aboverulesep=0ex 
\belowrulesep=0ex 
\caption{Comparison between first (Latency(V1) $\leq$) and last (Latency(V3) $\leq$) buckets across various model characteristics.}
\label{table:bucketconv}
\begin{tabular}{l|l|l|}
\cline{2-3}
&\textbf{Latency(V1) $\leq$}&\textbf{Latency(V3) $\leq$}\\\midrule
\multicolumn{1}{|l|}{Avg. $\#$ of Conv~3$\times$3}&1.53&0.78\\\midrule
\multicolumn{1}{|l|}{Avg. $\#$ of Conv~1$\times$1}&1.65&2.17\\\midrule
\multicolumn{1}{|l|}{Avg. $\#$ of MaxPool~3$\times$3}&1.66&1.77\\\midrule
\multicolumn{1}{|l|}{Avg. Graph Depth}&4.96&4.64\\\midrule
\multicolumn{1}{|l|}{Avg. $\#$ of Trainable Parameters}&7,054,471.34&1,417,485.36\\\midrule
\end{tabular}
\end{table}
\begin{figure}[t]
    \centering
    \subfloat[NASBench Cell]{
    \includegraphics[width=0.18\textwidth]{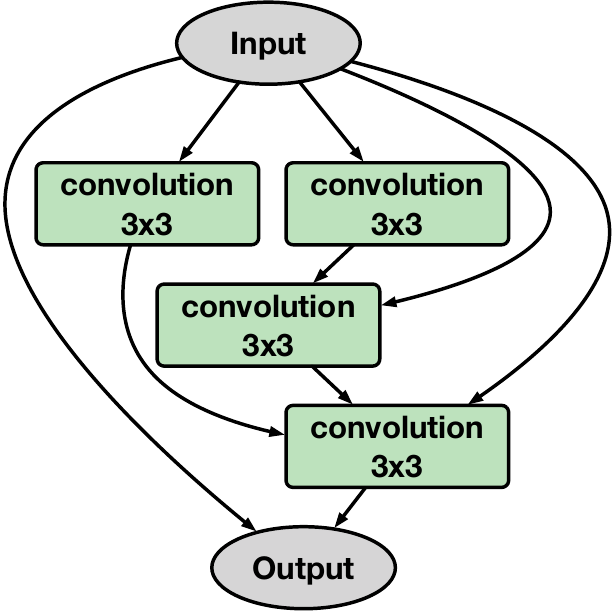}
    \label{fig:best_acc_graph}}
    \subfloat[Latency]{
    \scalebox{0.8}{
    \begin{tabular}[b]{lc}\hline
      \textbf{Accelerator} & \textbf{Latency} \\ \hline
      V1&4.633768\\
      V2&4.185697\\
      V3&4.535305\\ \hline
    \end{tabular}
    }
    \label{fig:best_acc_results}}
    \caption{(a) NASBench cell with highest mean validation accuracy after 108 epochs (95.055$\%$) and (b) the latency of running the NASBench cell on various Edge TPU accelerators. The total number of parameters of the convolutional neural network built from this NASBench cell is  41,557,898. For the highest accuracy model, V2 yields the lowest latency.} 
    \label{fig:best_acc}
\end{figure}
\begin{figure}[t]
    \centering
    \subfloat[NASBench Cell]{
    \includegraphics[width=0.18\textwidth]{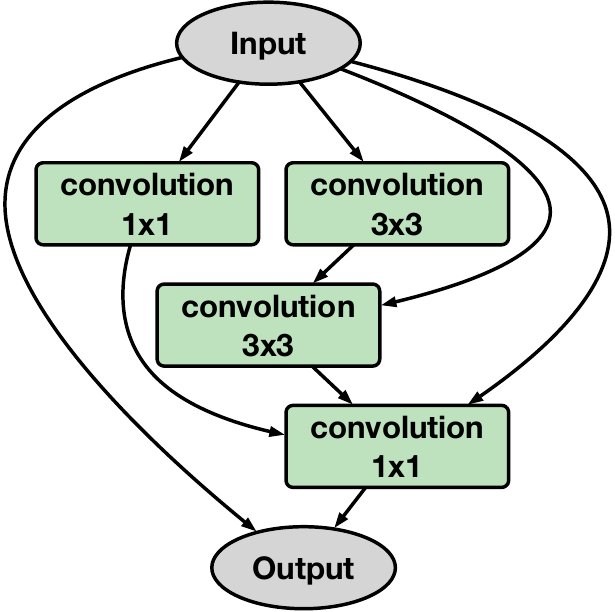}
    \label{fig:second_best_acc_graph}}
    \subfloat[Latency]{
    \scalebox{0.8}{
    \begin{tabular}[b]{lc}\hline
      \textbf{Accelerator} & \textbf{Latency (Speedup)} \\ \hline
      V1&2.597874~(1.78$\times$)\\
      V2&2.679829~(1.56$\times$)\\
      V3&2.799071~(1.62$\times$)\\ \hline
    \end{tabular}
    }
    \label{fig:second_best_acc_results}}
    \caption{(a) NASBench cell with the \textit{second} highest mean validation accuracy after 108 epochs (94.895$\%$) and (b) the latency of running the NASBench cell on various Edge TPU accelerators. The values in the parentheses represent the speedup over the NASBench cell with highest mean validation accuracy (Figure~\ref{fig:best_acc}).  The total number of parameters of the convolutional neural network built from this NASBench cell is 25,042,826 (66$\%$ less number of parameters compared to Figure~\ref{fig:best_acc}). For the highest accuracy of NASBench models, V1 yields the lowest latency and maximum speedup.} 
    \label{fig:second_best_acc}
\end{figure}
\begin{figure}[t]
    \centering
    \includegraphics[width=0.46\textwidth]{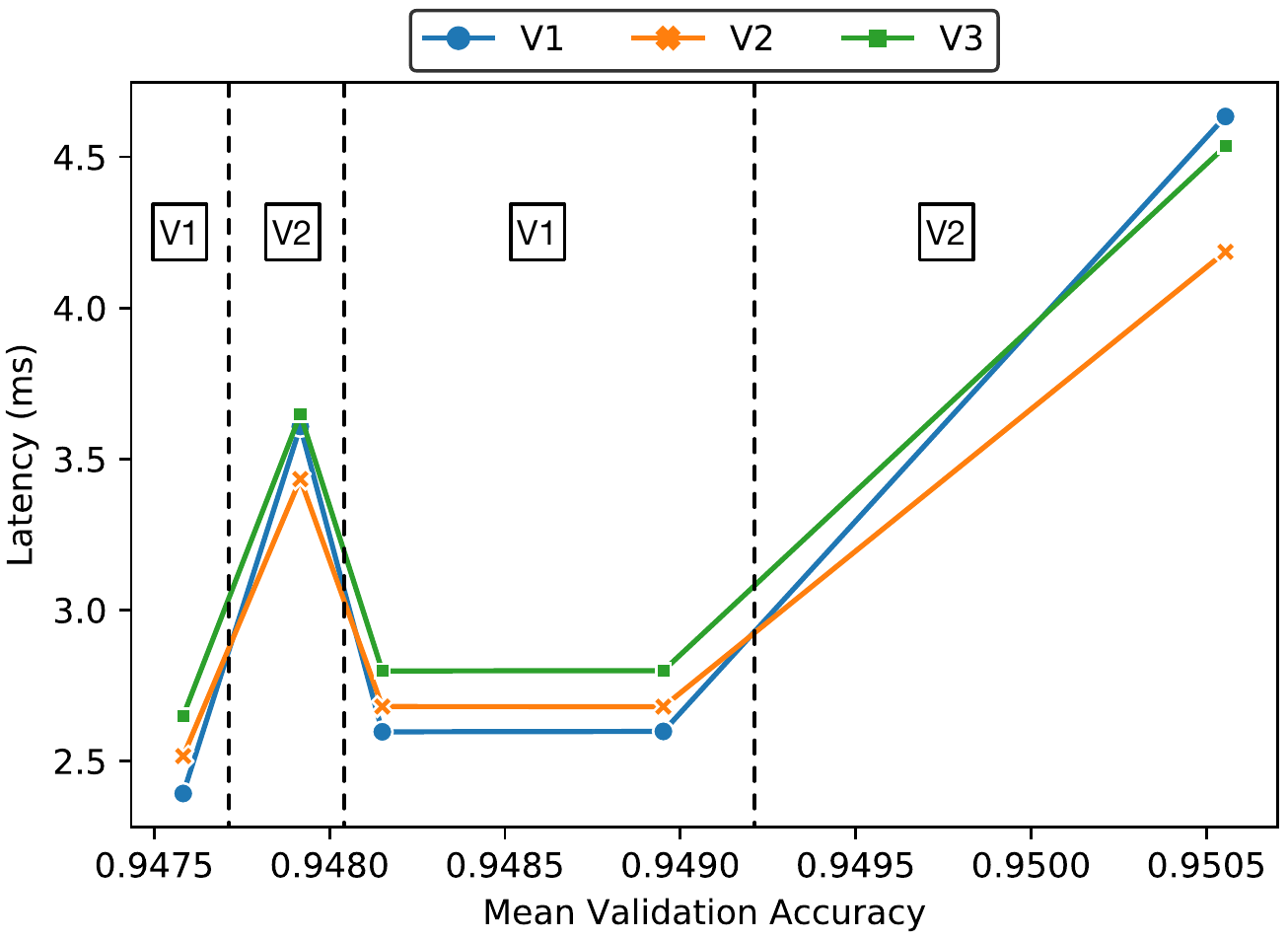}
    \caption{Comparing latency vs. mean validation accuracy for the top five performing models in terms of mean validation accuracy. Each section demarcated by dashed lines highlights the Edge TPU configuration yielding the lowest latency.}
    \label{fig:lat_acc_top5}
\end{figure}
\begin{figure}[!h]
    \centering
    \subfloat[Accuracy vs. Graph Depth]{
    \includegraphics[width=0.24\textwidth]{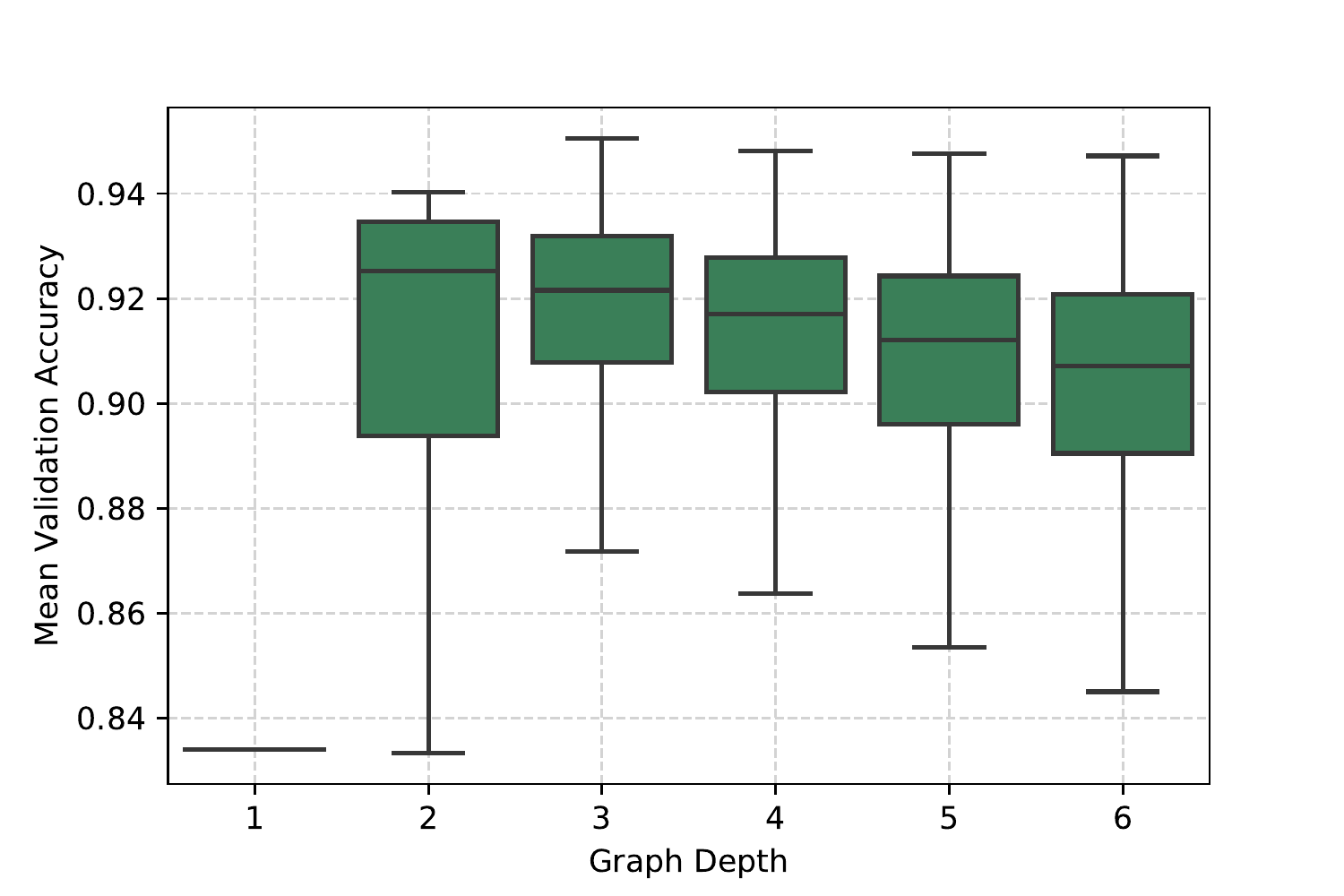}
    \label{fig:depth_val_acc}}
    \subfloat[Accuracy vs. Graph Width]{
    \includegraphics[width=0.24\textwidth]{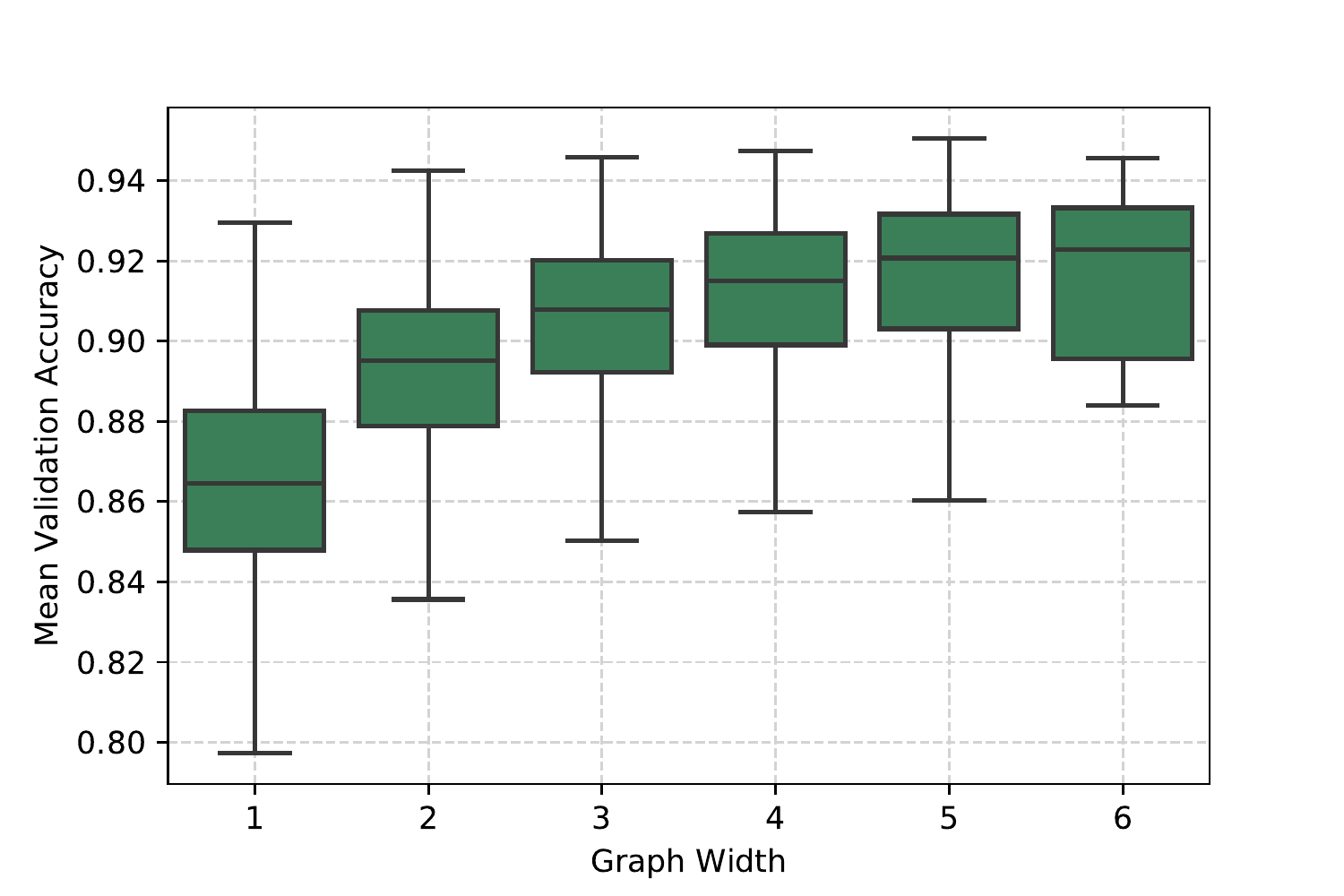}
    \label{fig:width_val_acc}}
    \caption{The comparison between mean validation accuracy vs. (a) graph depth and (b) graph width. Graph depth is measured by the length of longest path from input node to output node. Graph width is measured by maximum directed cut on the graph (same terminology as NASBench-101~\cite{nasbench101}).} 
    \label{fig:depth_width_acc}
\end{figure}

\niparagraph{Analysis of models with high accuracy.}
Figure~\ref{fig:best_acc_graph} illustrates the architecture of the NASBench cell that forms a convolutional neural model with the highest accuracy (95.055$\%$) in the NASBench dataset. 
Figure~\ref{fig:best_acc_results} annotates the neural architecture with the latency numbers across the three studied accelerators.
\bench{V2} is the winner across all the accelerator configurations with the latency of 4.18~ms (10$\%$ less than \bench{V1} accelerator).

To better understand the trade-off between accuracy and latency, we also study the NASBench neural model with the second highest mean validation accuracy, shown in Figure~\ref{fig:second_best_acc_graph}.
The main observation here is that 0.16$\%$ trade-off in accuracy (95.055~$\%\rightarrow$~94.895$\%$) yields up to 1.78$\times$ improvement in latency in \bench{V1}.
Note that, even though the accuracy degradation between these two neural models (Figure~\ref{fig:best_acc} and Figure~\ref{fig:second_best_acc}) are minimal, there are 66\% fewer parameters in the model shown in Figure~\ref{fig:second_best_acc}, a key contributing factor to lower latency.
Finally, it is interesting to observe that for the NASBench cell in Figure~\ref{fig:second_best_acc_graph} \bench{V1} yields lower latency, in contrast to Figure~\ref{fig:best_acc_graph} in which \bench{V2} performs better.
We can attribute this result to (a) the lower I/O bandwidth in \bench{V1} that contributes more to reducing latency for neural models with fewer number of parameters and (b) the higher efficiency of \bench{V1} in running 1$\times$1 convolution operations.

\niparagraph{Analysis of accelerator performance for high-accuracy convolutional models.}
Figure~\ref{fig:lat_acc_top5} illustrates the comparison between latency and mean validation accuracy for the top five models with highest accuracy.
The dashed lines in the figure split the curve into four regions showing which Edge TPU accelerator yields the lowest latency for that region.
As mentioned before, for the highest accuracy model, \bench{V2} delivers the lowest latency.
However, for the next two models, \bench{V1} is the winner.
One of the main reasons for this trend is the type of operations that are used in each region.
\bench{V1} generally performs better for the models with a larger number of 1$\times$1 convolutional operators.
Furthermore, this trend highlights the significant headroom for reducing the accelerator latency by slightly compromising the neural model accuracy. 
\begin{figure*}[th]
    \centering
    \subfloat[V1 (Latency vs. Graph Depth)]{
    \includegraphics[width=0.33\textwidth]{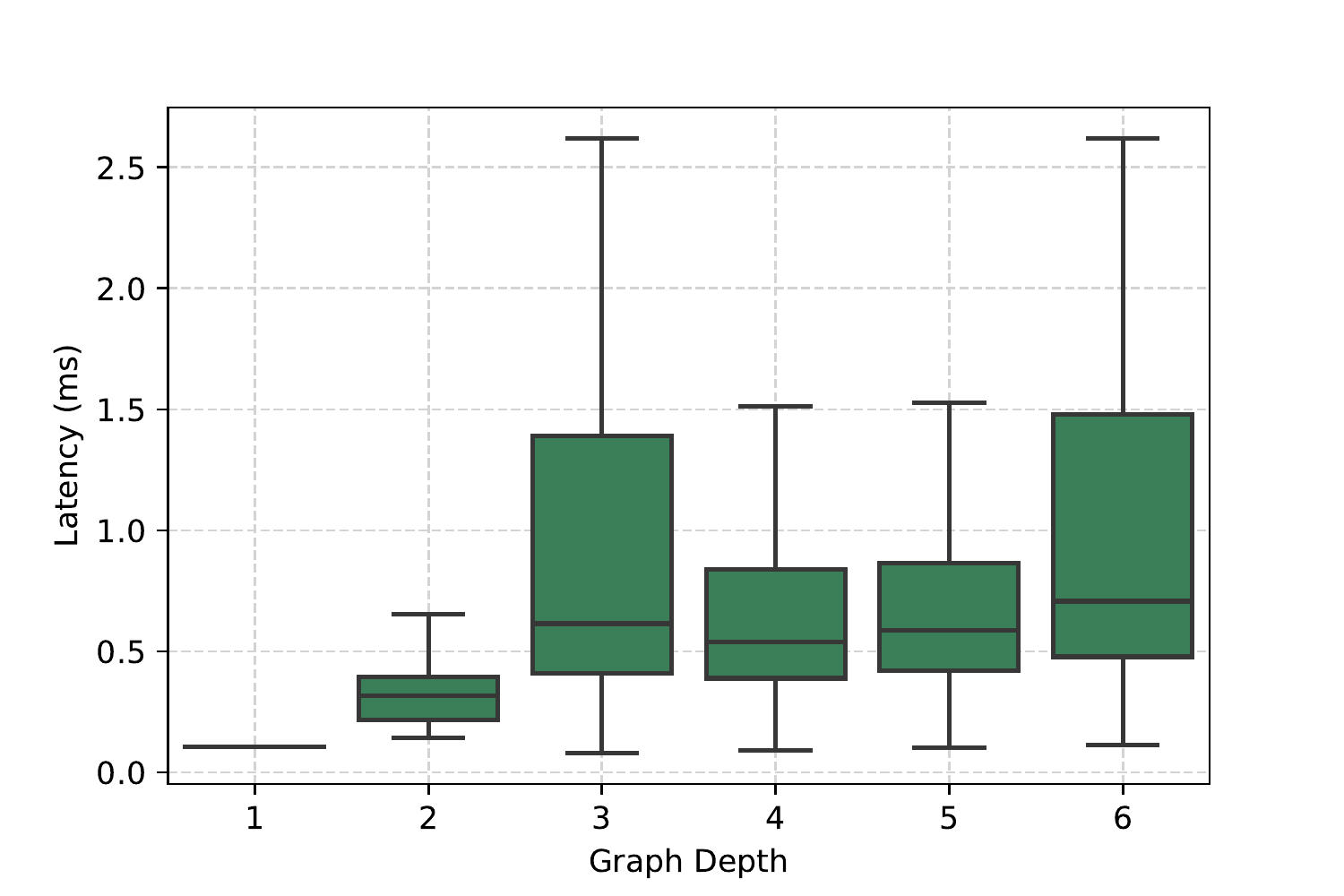}
    \label{fig:depth_dc_lat}}
    \subfloat[V2 (Latency vs. Graph Depth)]{
    \includegraphics[width=0.33\textwidth]{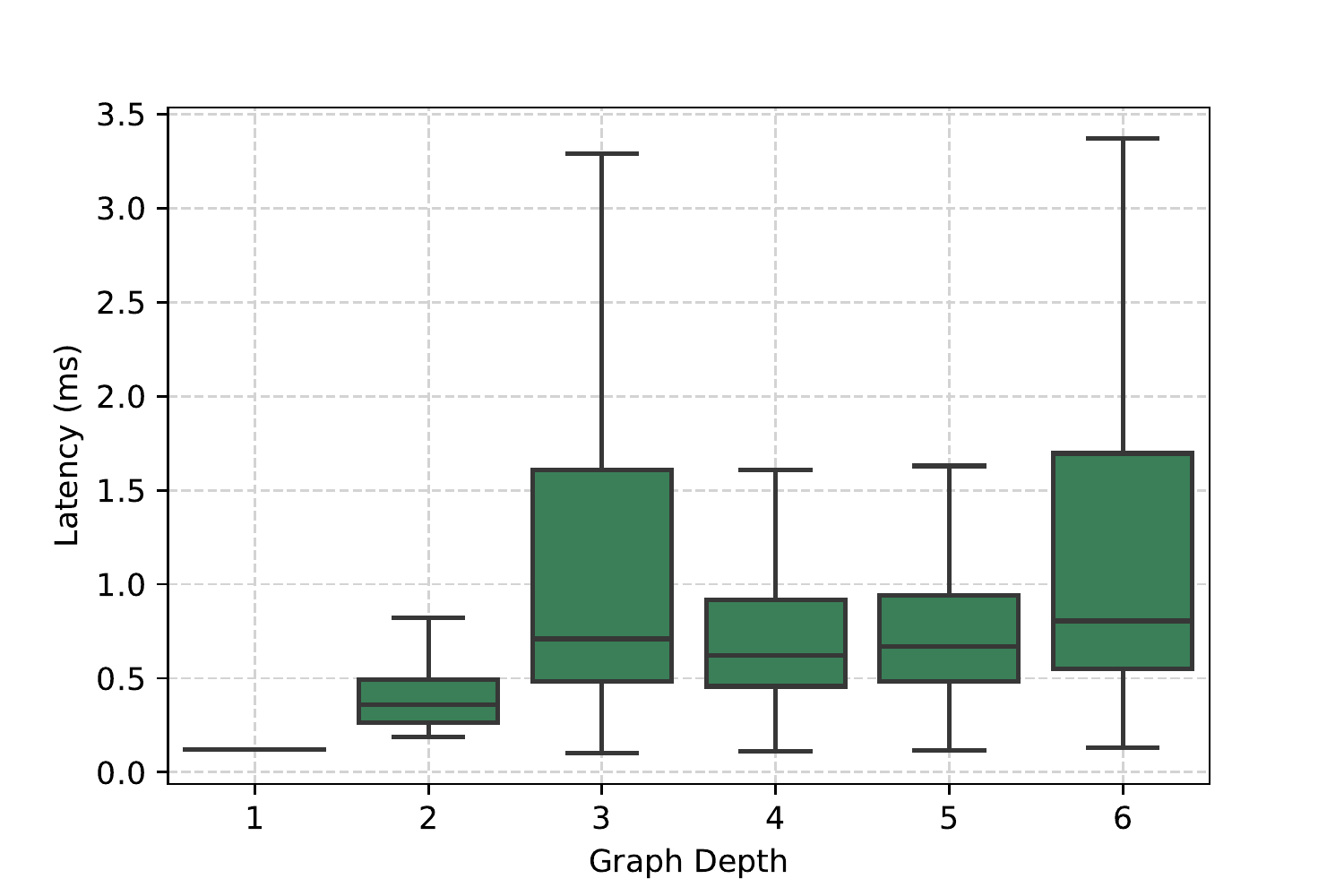}
    \label{fig:depth_msoc1_lat}}
    \subfloat[V3 (Latency vs. Graph Depth)]{
    \includegraphics[width=0.33\textwidth]{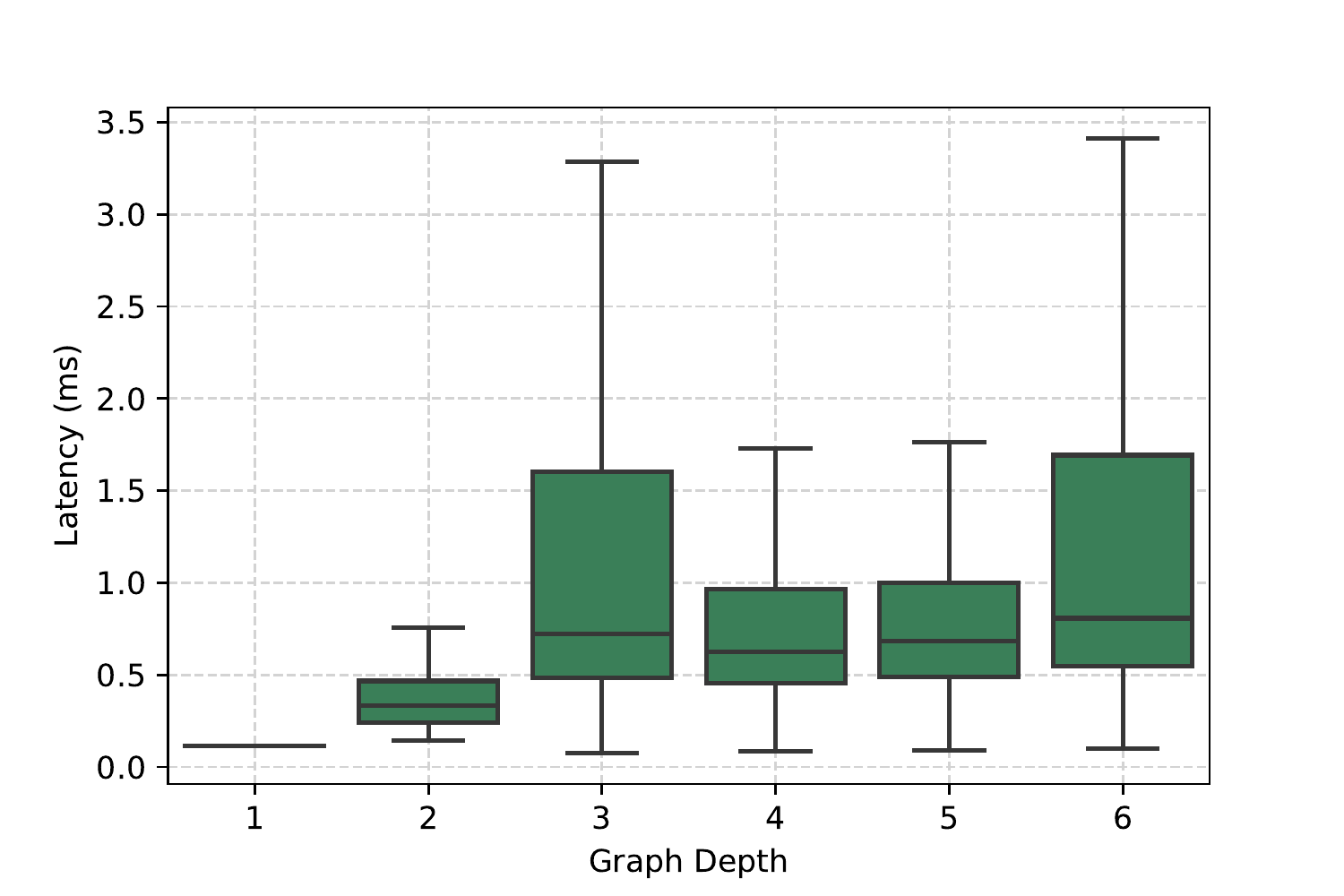}
    \label{fig:depth_msoc2_lat}}\\
    \subfloat[V1 (Latency vs. Graph Width)]{
    \includegraphics[width=0.33\textwidth]{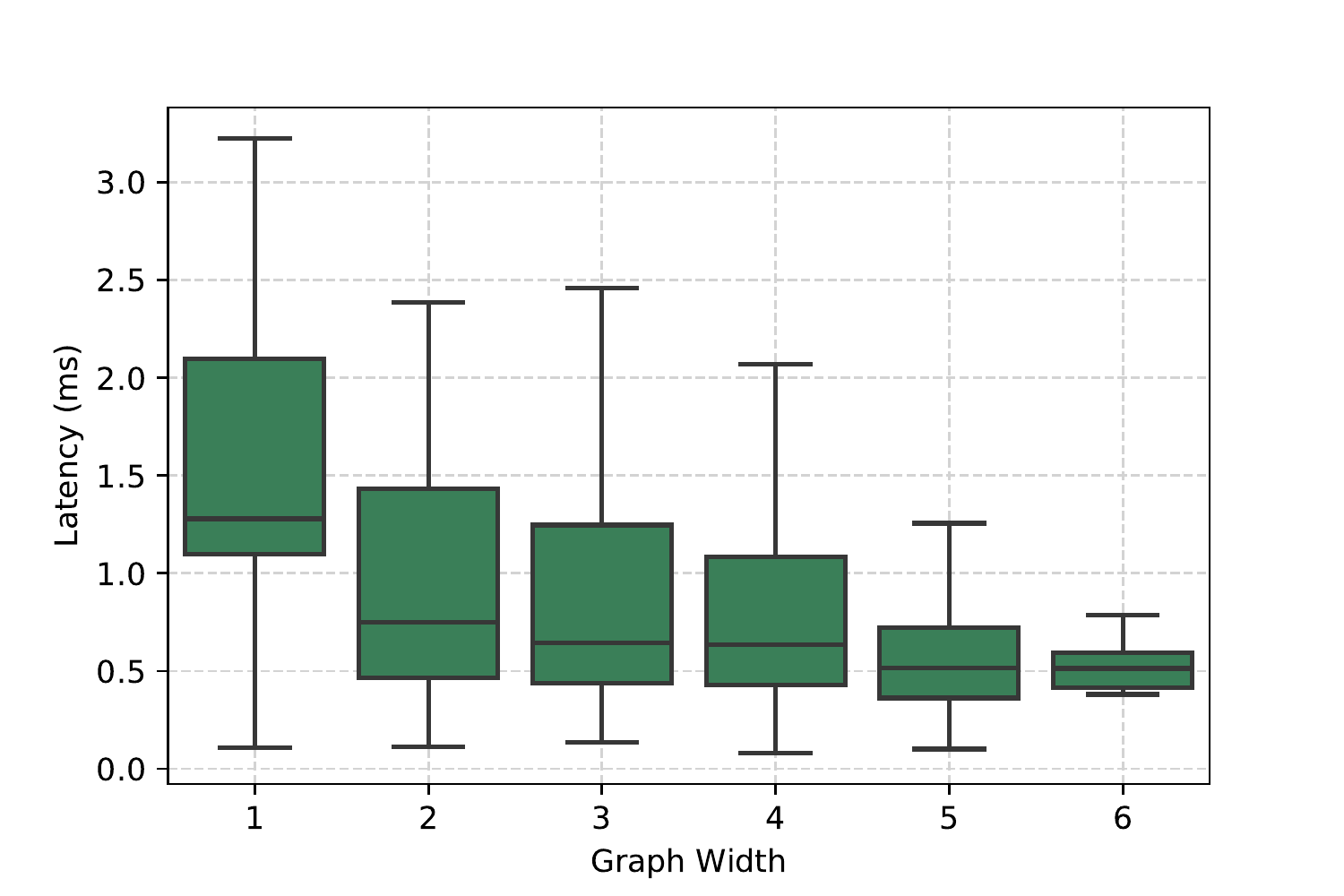}
    \label{fig:width_dc_lat}}
    \subfloat[V2 (Latency vs. Graph Width)]{
    \includegraphics[width=0.33\textwidth]{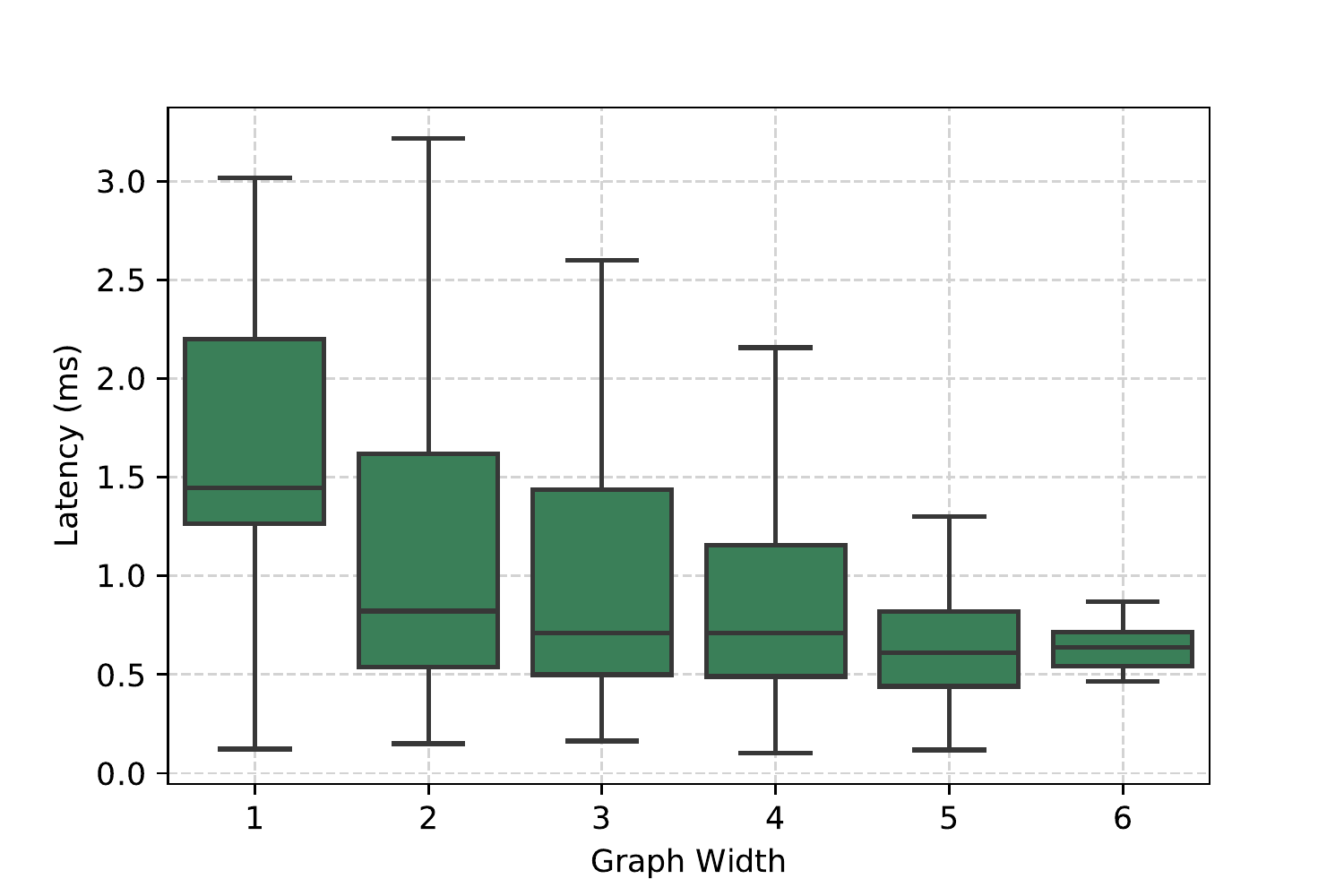}
    \label{fig:width_msoc1_lat}}
    \subfloat[V3 (Latency vs. Graph Width)]{
    \includegraphics[width=0.33\textwidth]{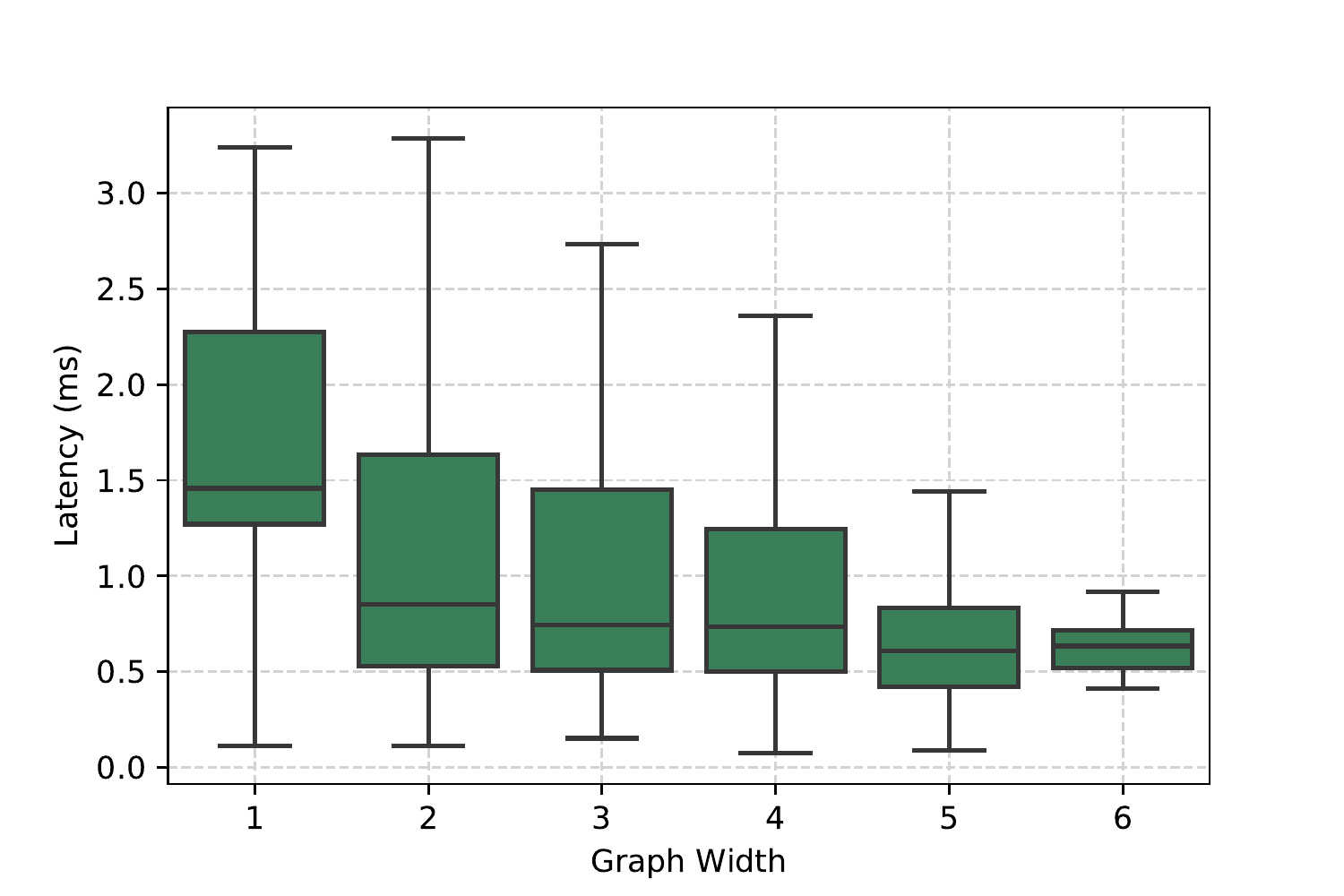}
    \label{fig:width_msoc2_lat}}\\
    \caption{The comparison between latency and graph depth (first row) and width (second row) across three Edge TPU configurations. Graph depth is measured by the length of longest path from input node to output node. Graph width is measured by maximum directed cut on the graph (same terminology as NASBench-101~\cite{nasbench101}).} 
    \label{fig:depth_width_lat}
\end{figure*}
\begin{figure*}[th]
    \centering
    \subfloat[V1]{
    \includegraphics[width=0.32\textwidth]{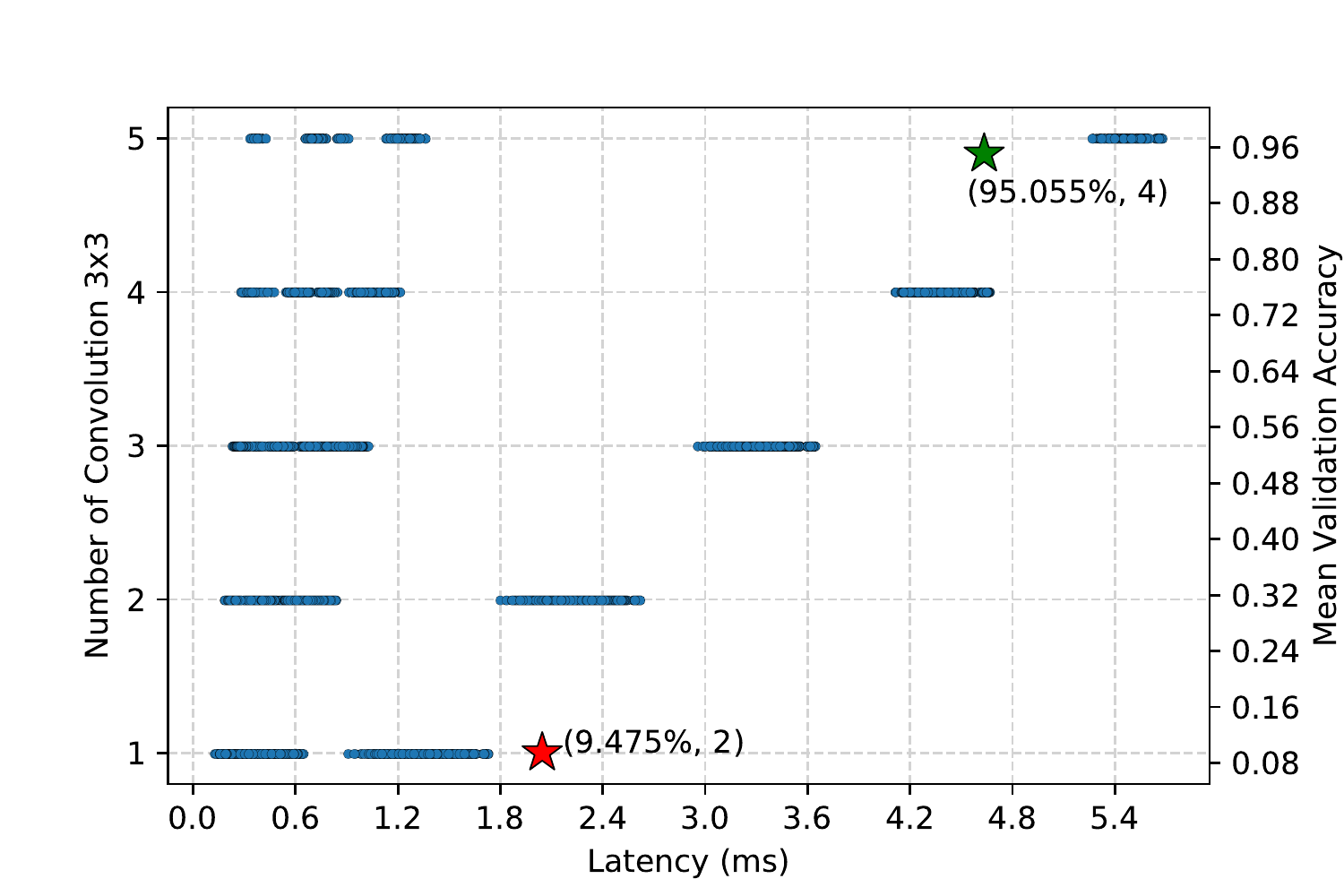}
    \label{fig:dc_num_conv3x3}}
    \subfloat[V2]{
    \includegraphics[width=0.32\textwidth]{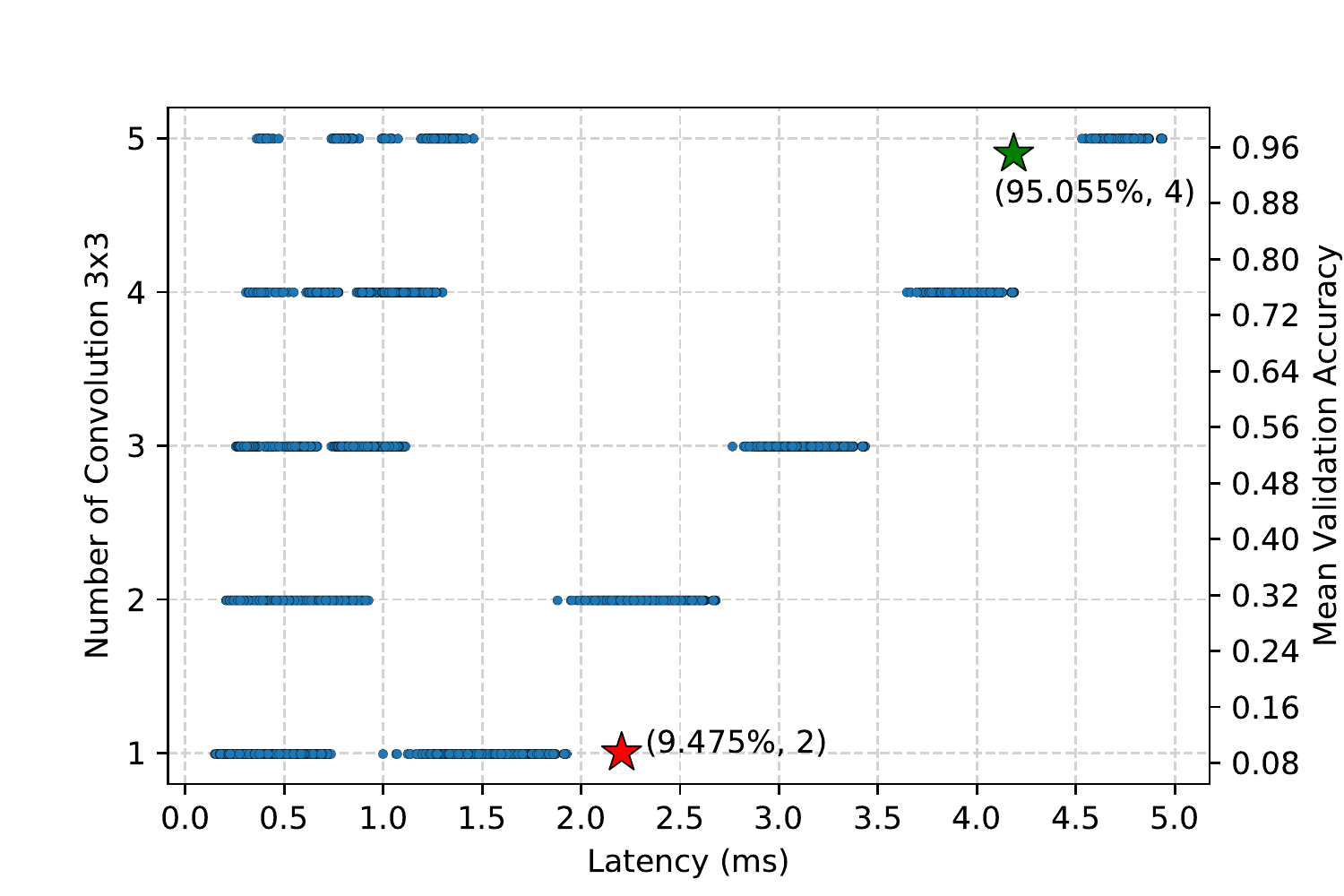}
    \label{fig:msoc1_num_conv3x3}}
    \subfloat[V3]{
    \includegraphics[width=0.32\textwidth]{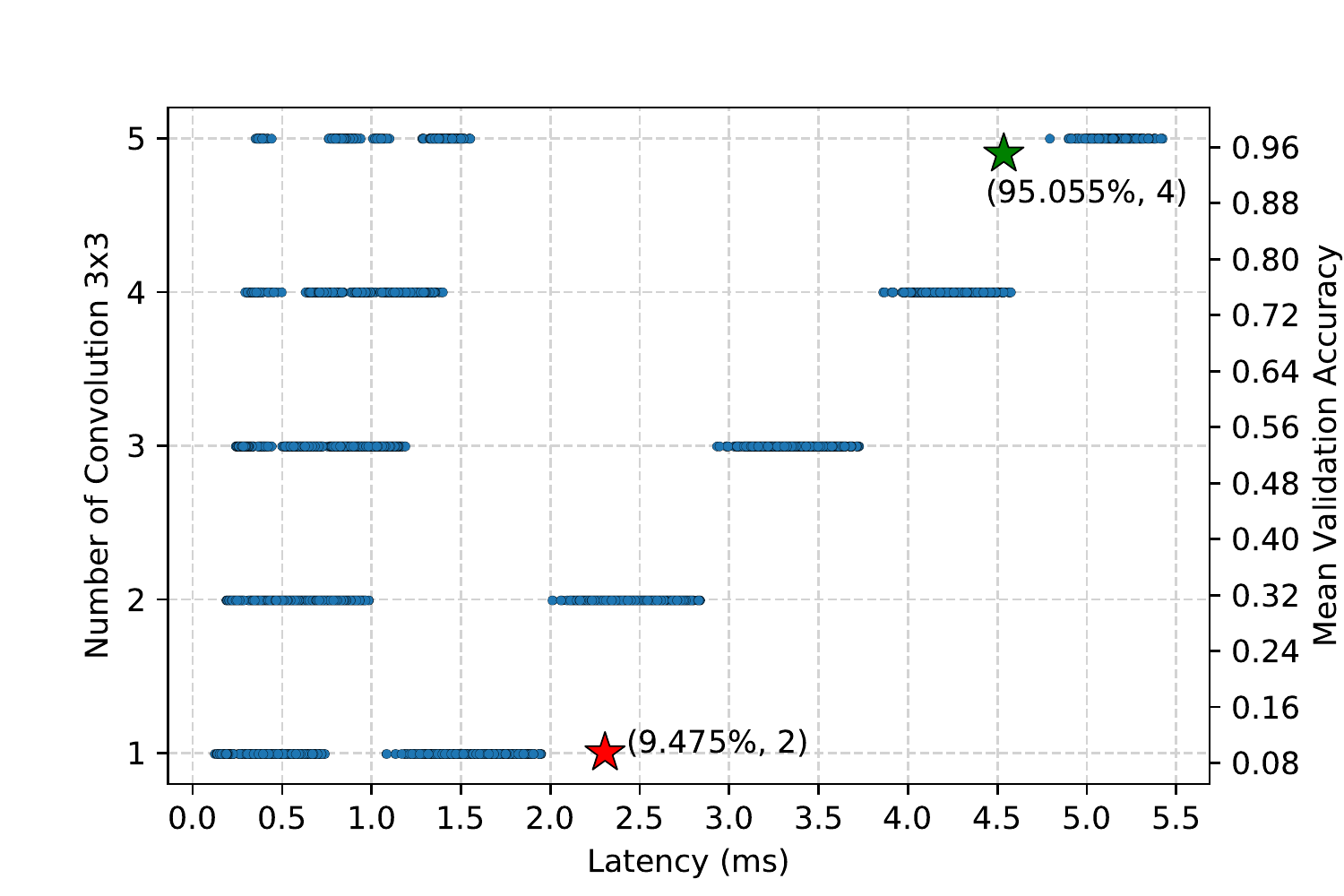}
    \label{fig:msoc2_num_conv3x3}}\\
    \subfloat[V1]{
    \includegraphics[width=0.32\textwidth]{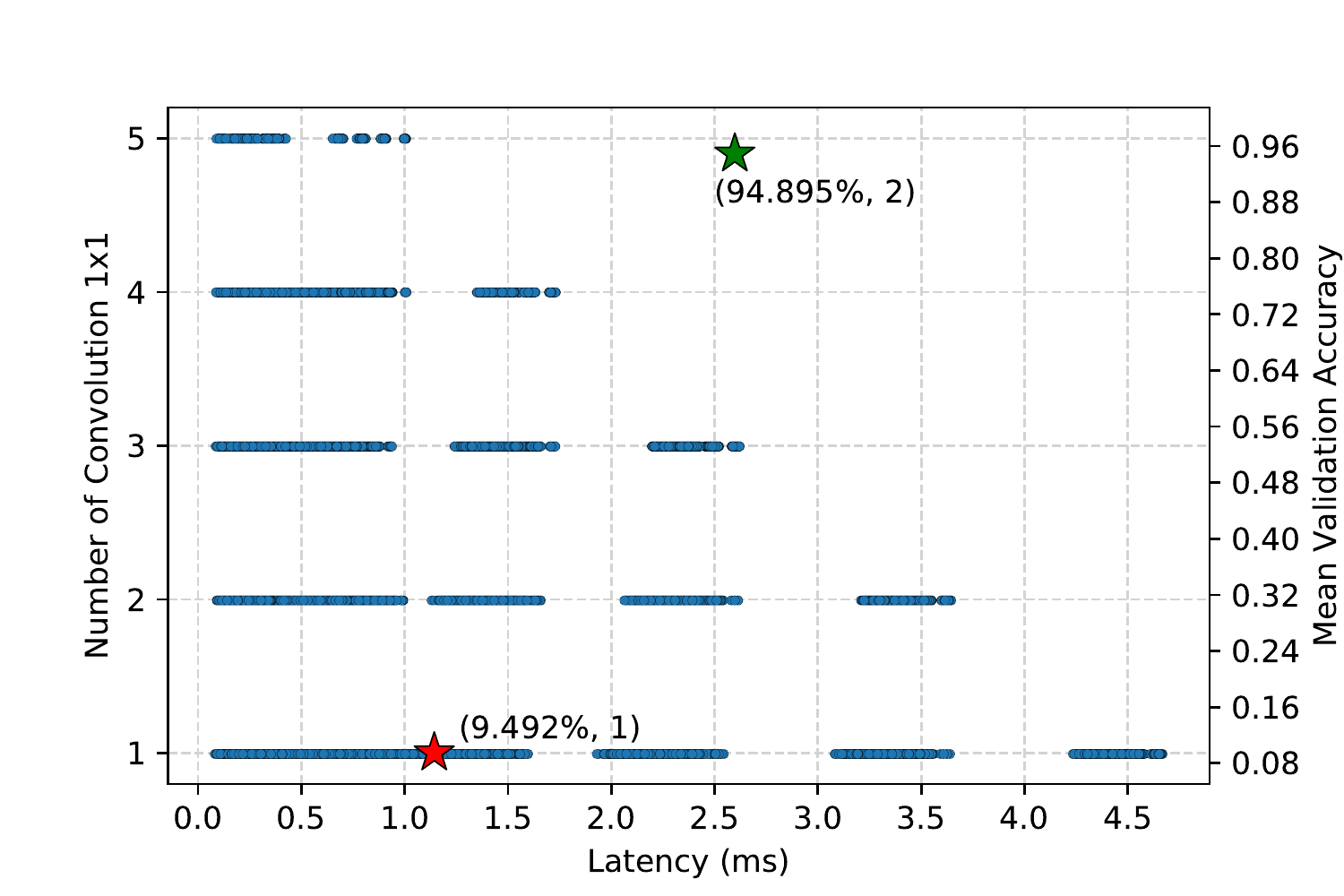}
    \label{fig:dc_num_conv1x1}}
    \subfloat[V2]{
    \includegraphics[width=0.32\textwidth]{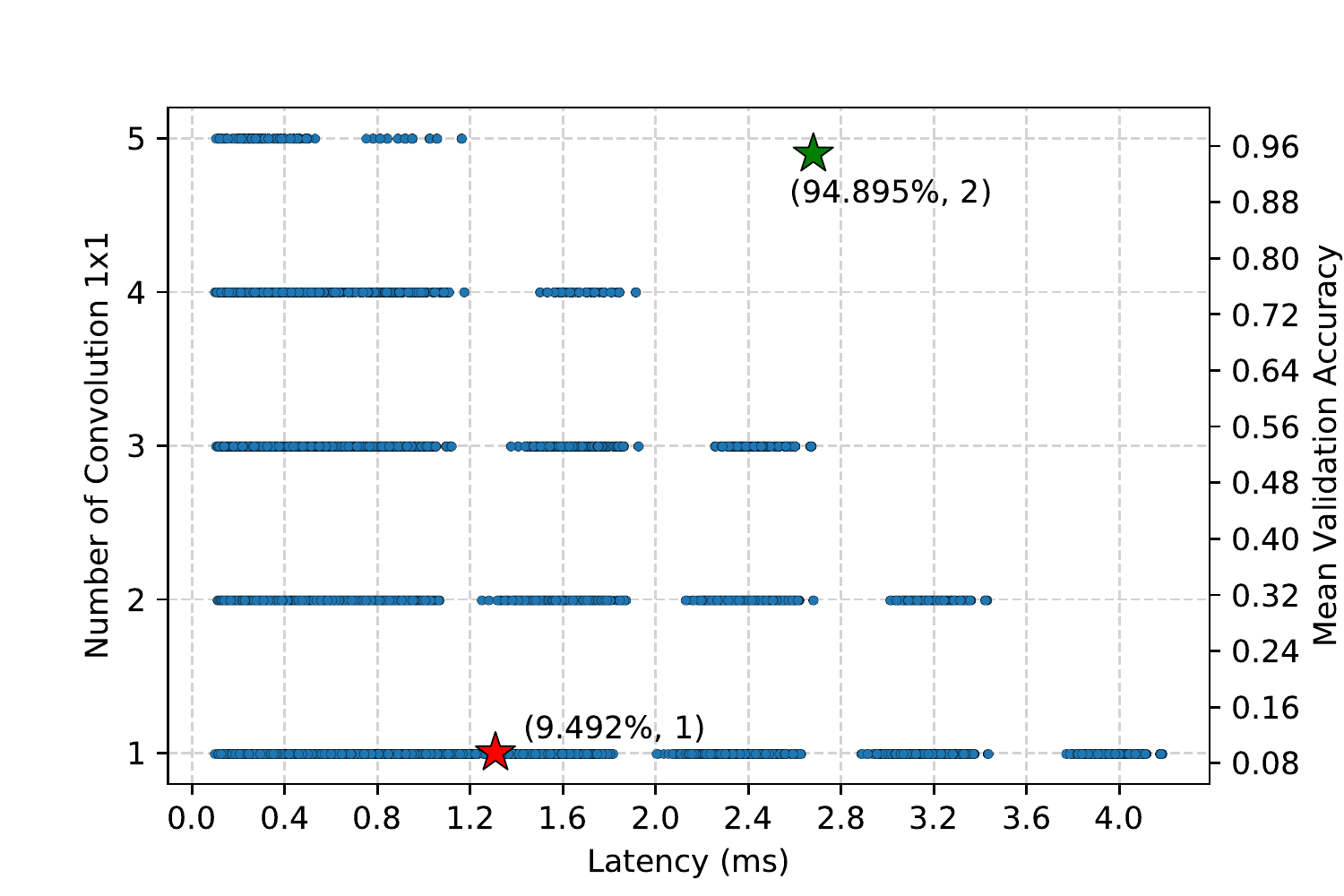}
    \label{fig:msoc1_num_conv1x1}}
    \subfloat[V3]{
    \includegraphics[width=0.32\textwidth]{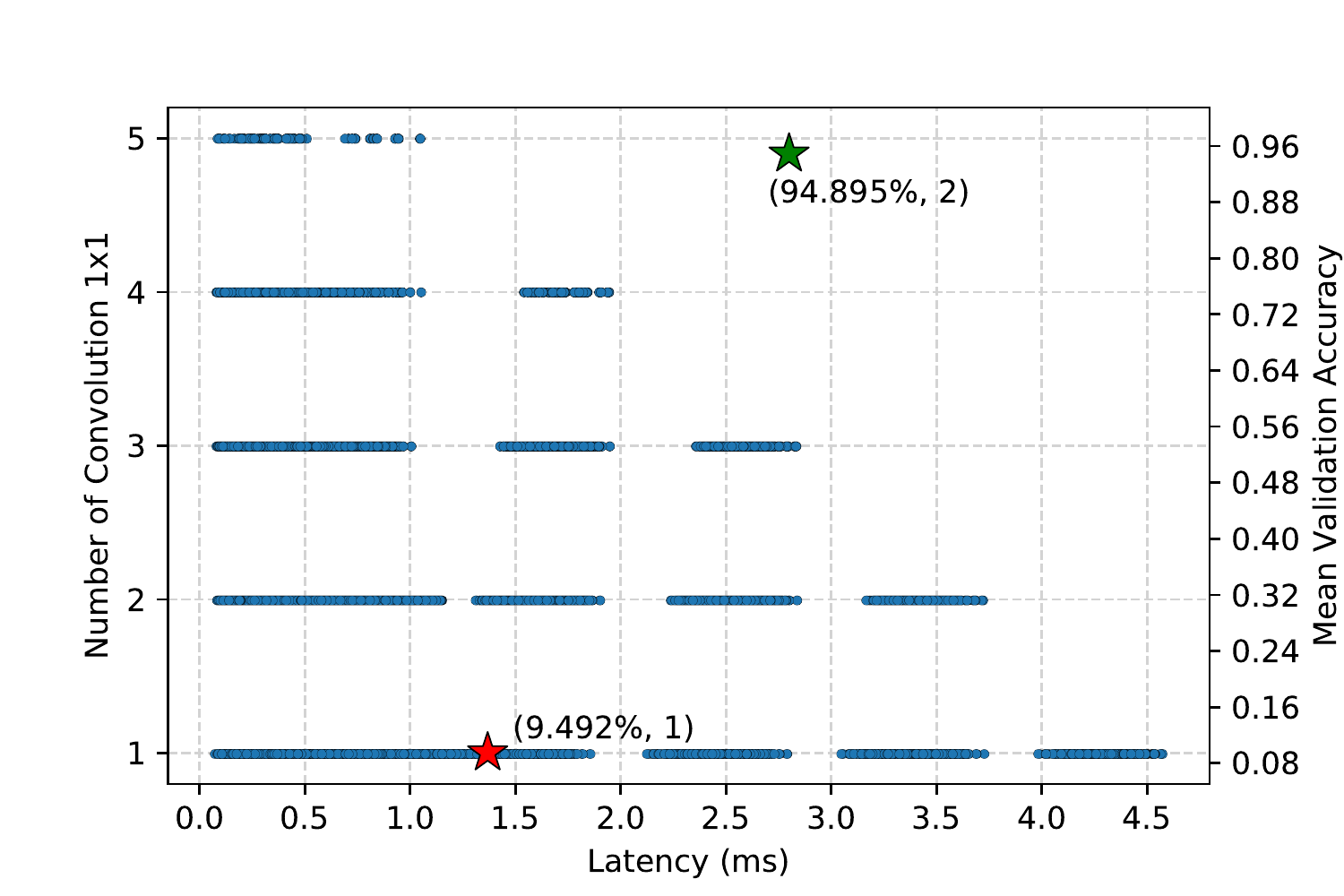}
    \label{fig:msoc2_num_conv1x1}}
    \\
    \subfloat[V1]{
    \includegraphics[width=0.32\textwidth]{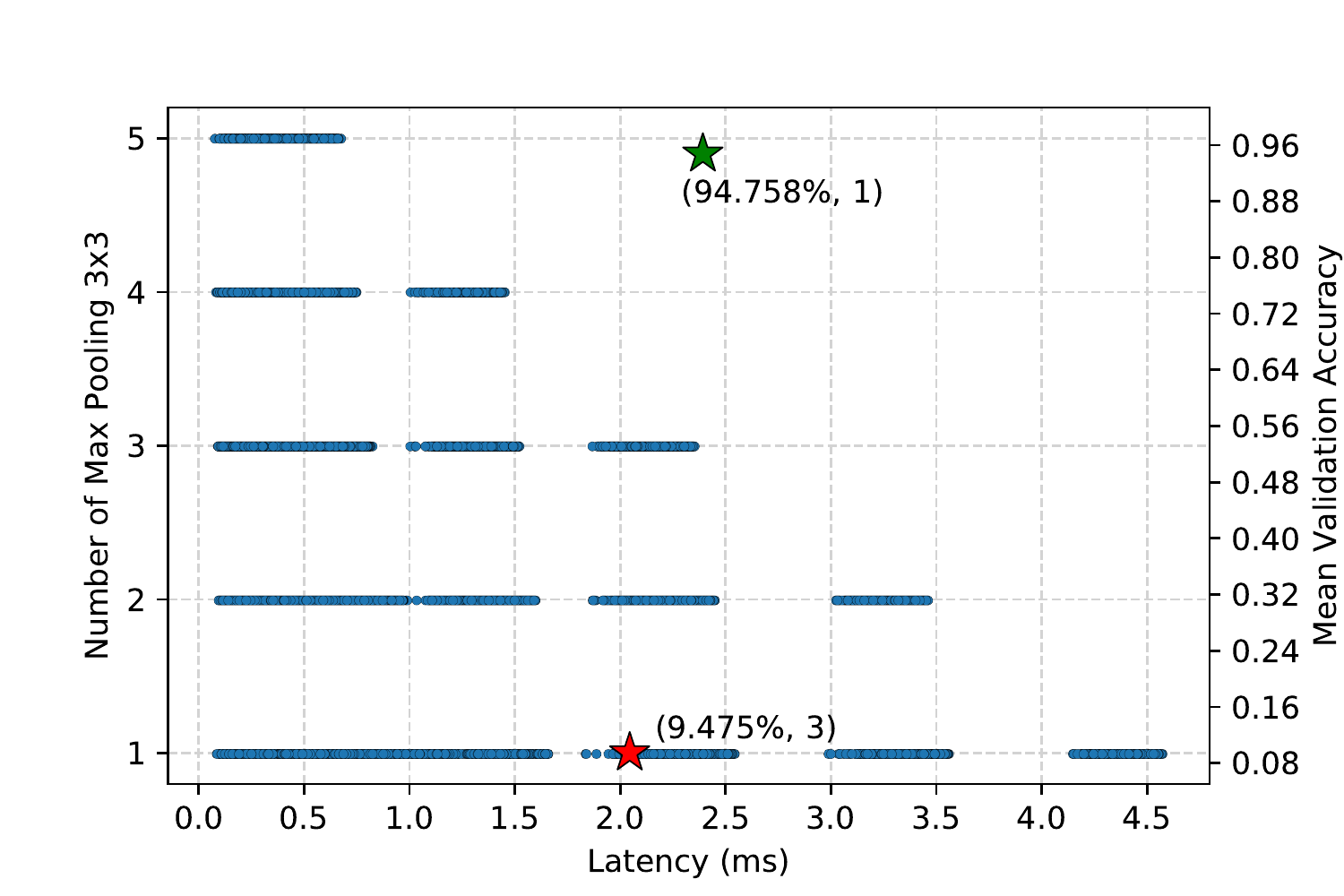}
    \label{fig:dc_num_maxpooling3x3}}
    \subfloat[V2]{
    \includegraphics[width=0.32\textwidth]{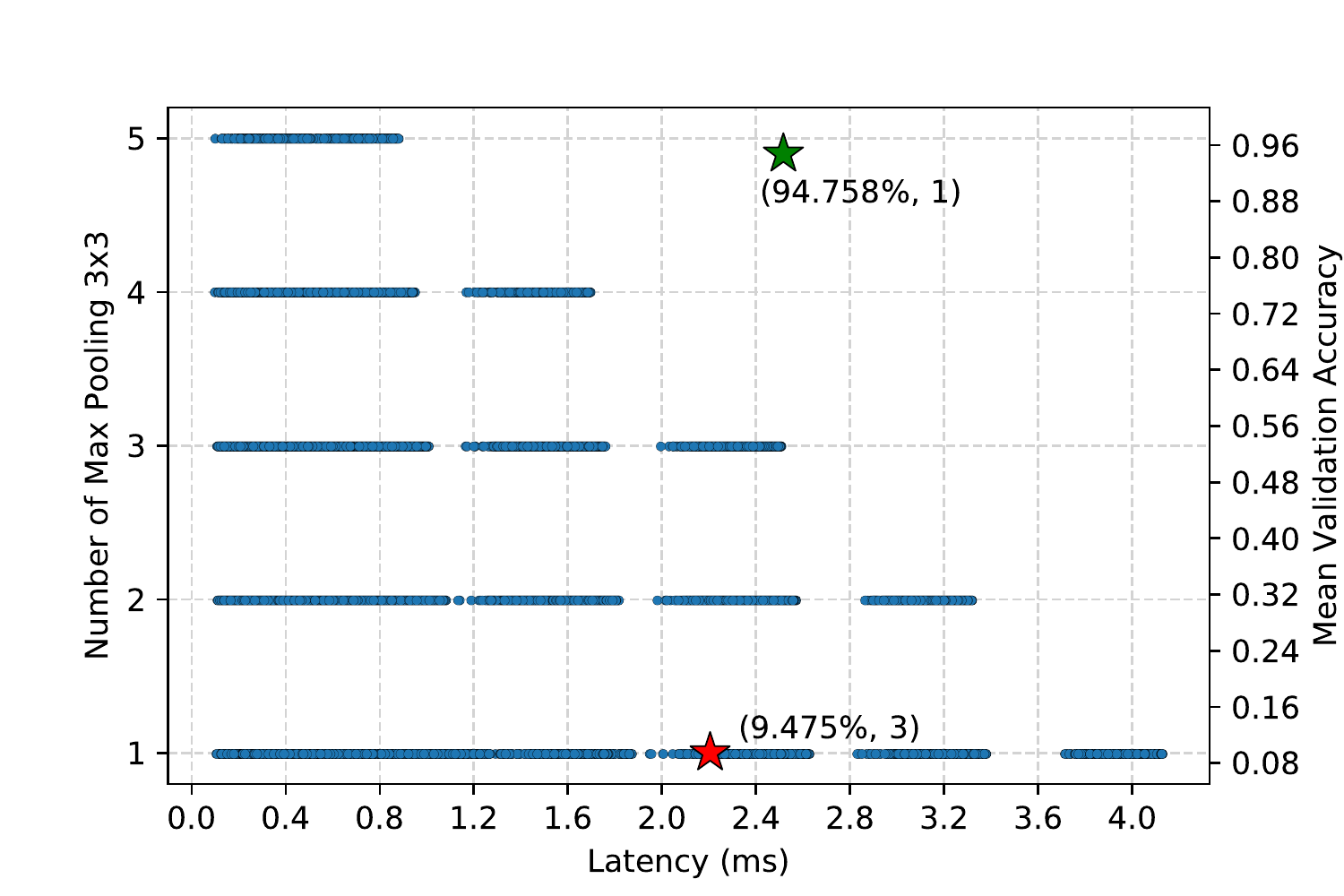}
    \label{fig:msoc1_num_maxpooling3x3}}
    \subfloat[V3]{
    \includegraphics[width=0.32\textwidth]{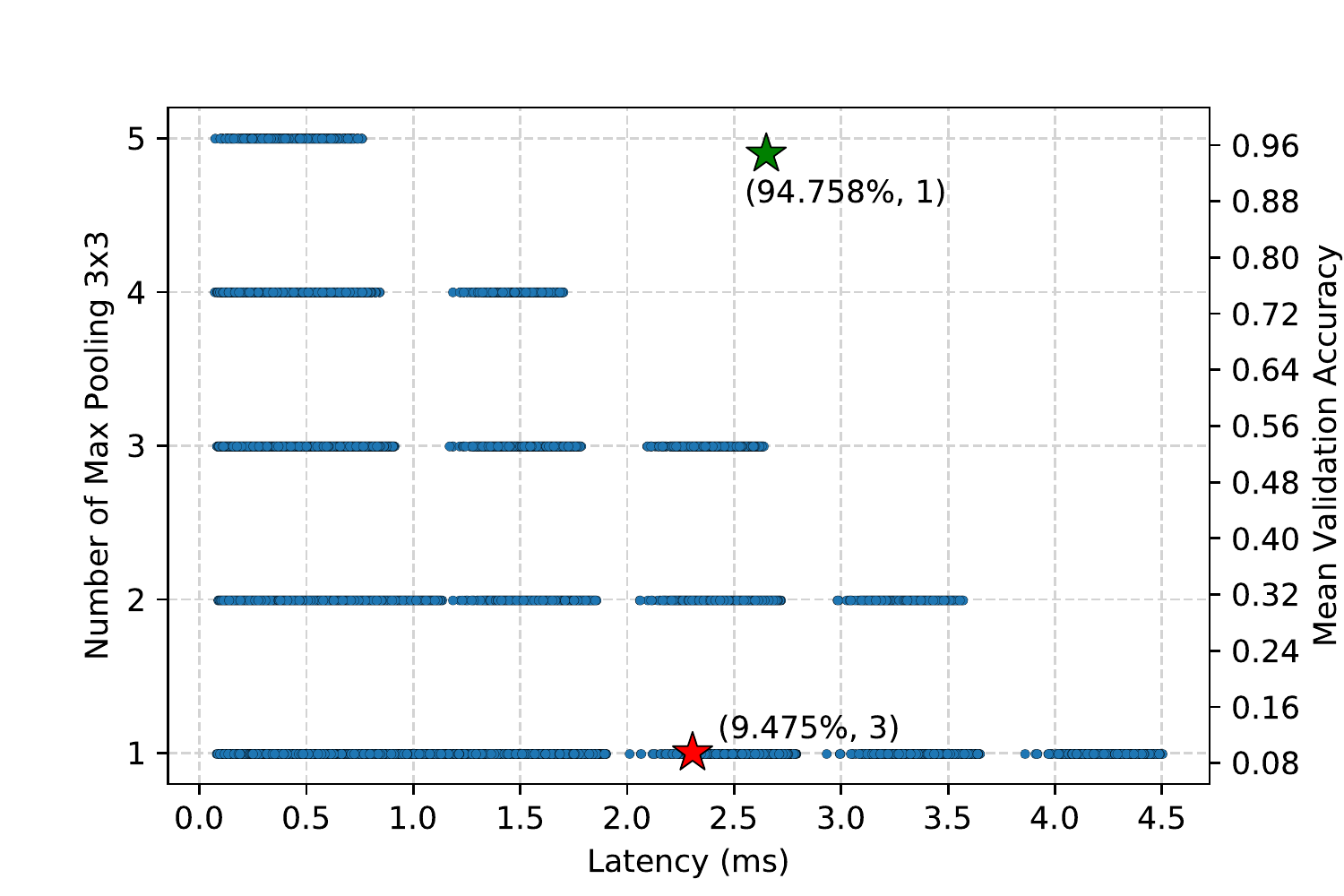}
    \label{fig:msoc2_num_maxpooling3x3}}
    \caption{The scatter plot showing the number of valid operations, namely convolution 3$\times$3 (first row; a-c), (b) convolution 1$\times$1 (second row; d-f), and max-pooling 3$\times$3 (third row; g-i)) for all the convolutional models in NASBench-101~\cite{nasbench101} vs. the measured inference latency on three difference configurations of the Google Edge TPU accelerators. 
    The green and red stars in each operation category indicate the maximum and minimum mean validation accuracy vs. latency, respectively. The highlighted values in parenthesis are (mean validation accuracy, the number of operations in the corresponding operation category).  
    This figure is best viewed in color.} 
    \label{fig:op_lat}
\end{figure*}
\begin{figure}[t]
    \centering
    \includegraphics[width=0.49\textwidth]{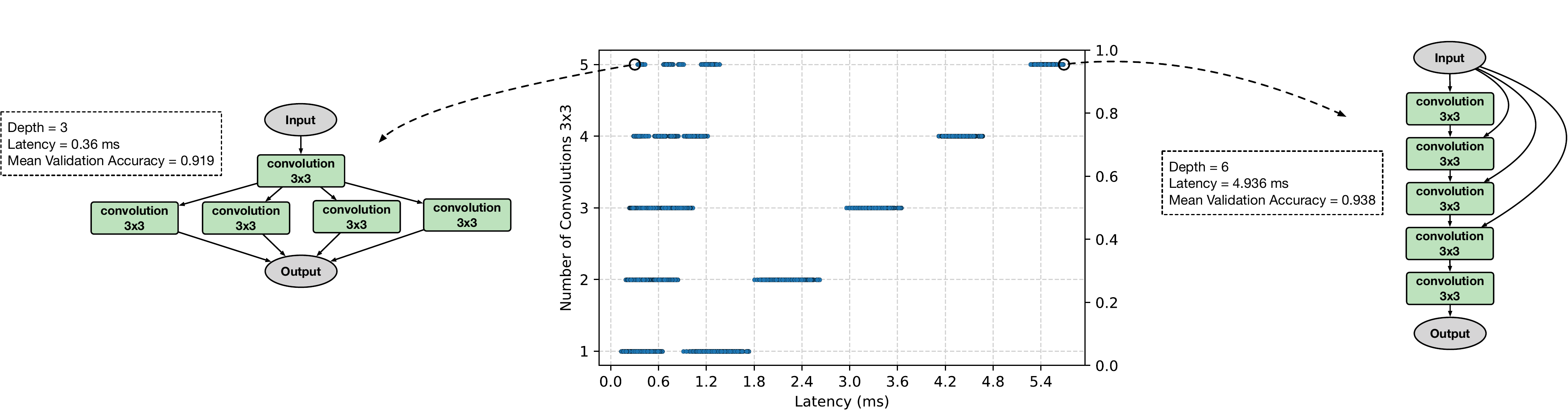}
    \caption{Scatter plot showing the number of convolution 3x3 operations in a NASBench cell vs. the measured inference latency including the NASBench cells that have 5 3$\times$3 convolutions each, and have the lowest and highest inference latency on the V2 configuration.}
    \label{fig:conv3x3_with_models}
\end{figure}

\niparagraph{Neural model architecture impact on accelerator performance.}
First, we investigate the role of graph depth (Figure~\ref{fig:depth_val_acc}) and graph width (Figure~\ref{fig:width_val_acc}) in terms of mean validation accuracy\footnote{Similar methodology used in NASBench-101~\cite{nasbench101}.}.
The whiskers in Figure~\ref{fig:depth_width_acc} show that depth of three and width of five are optimal in terms of validation accuracy.
The results also show that increasing depth beyond three has a negative impact on the accuracy of the model.

We also study the impact of graph structure on the accelerator latency across three Edge TPU configurations (Figure~\ref{fig:depth_width_lat}).
The overall trend across all the accelerators shows that increasing graph depth increases latency, as the length of dependencies between operations increases.
However, this trend breaks (latency decreases) for graph depth of four and five.
On of the reason for such behavior is that the average number of model parameters for these graphs are lower (See Table~\ref{tab:depthparam}) than other graph depths.
\begin{table}
\scriptsize
\centering
\aboverulesep=0ex 
\belowrulesep=0ex 
\caption{Average number of model parameters vs. graph depth.}
\label{tab:depthparam}
\begin{tabular}{l|l}
\midrule
\textbf{Graph Depth}&\textbf{Avg. $\#$ of Parameters}\\\midrule
3&7,442,469.77\\\midrule
4&6,144,266.36\\\midrule
5&6,399,201.72\\\midrule
6&8,428,092.52\\\midrule
\end{tabular}
\end{table}
However, increasing graph width generally results in lower latency, as it improves the parallelism between operations.
Taking all of these results (graph structure, accuracy, and latency) into account, there is an interesting trade-off between model accuracy, accelerator performance, and graph structure.
The results show that increasing the graph depth beyond a limit (three in our dataset) does not improve the model accuracy.
However, increasing the graph width not only improves the model accuracy, but also tends to be more favorable in reducing the accelerator latency, mainly due to the higher parallelism between neural network operations.

\niparagraph{Correlation between latency and number of neural operations.}
To better understand the correlation between model complexity and accelerator performance, we further investigate the impact of operation types and number of trainable parameters on the accelerator latency.
Figure~\ref{fig:op_lat} shows the impact of number of 3$\times$3 convolution (first row), 1$\times$1 convolution (second row), and 3$\times$3 max pooling (third row) operations in each NASBench cell on inference latency across the studied accelerators.
We also annotate each figure with the maximum (green star marker) and minimum (red star marker) achievable model accuracy for each category of operations. The highlighted values in parenthesis are (mean validation accuracy, the number of operations in the corresponding operation category).
For example, Figure~\ref{fig:dc_num_conv3x3} shows that for the category of 3$\times$3 convolution operations the maximum achievable model accuracy is 95.055$\%$ with four 3$\times$3 convolution operation in the NASBench cell.

Due to the strict limit on the number of operations in each NASBench cell (seven), the latency increases as we increase the number of 3$\times$3 convolution operations.
This is because convolution 3$\times$3 has higher number of parameters compared to other operations (e.g. convolution 1$\times$1 and MaxPool 3$\times$3). Hence, convolution 3$\times$3 impacts latency more. In NASBench dataset, fewer convolution 1$\times$1 / MaxPool 3$\times$3 operations generally leads to more convolution 3$\times$3 (more parameters). 
However, for the same number of 3$\times$3 convolution operations in a NASBench cell, the latency numbers range from 0.2 ms to 5 ms.
Figure~\ref{fig:conv3x3_with_models} shows an example of two NASBench cell, one for each case.
For the cases where the number of 3$\times$3 convolution operations is high but the model depth is low, the inference latency will be low.
If the number of 3$\times$3 convolution operations and depth increase, the inference latency increases significantly.

\niparagraph{Correlation between latency and trainable parameters.}
\begin{figure}[!h]
    \centering
    \includegraphics[width=0.47\textwidth]{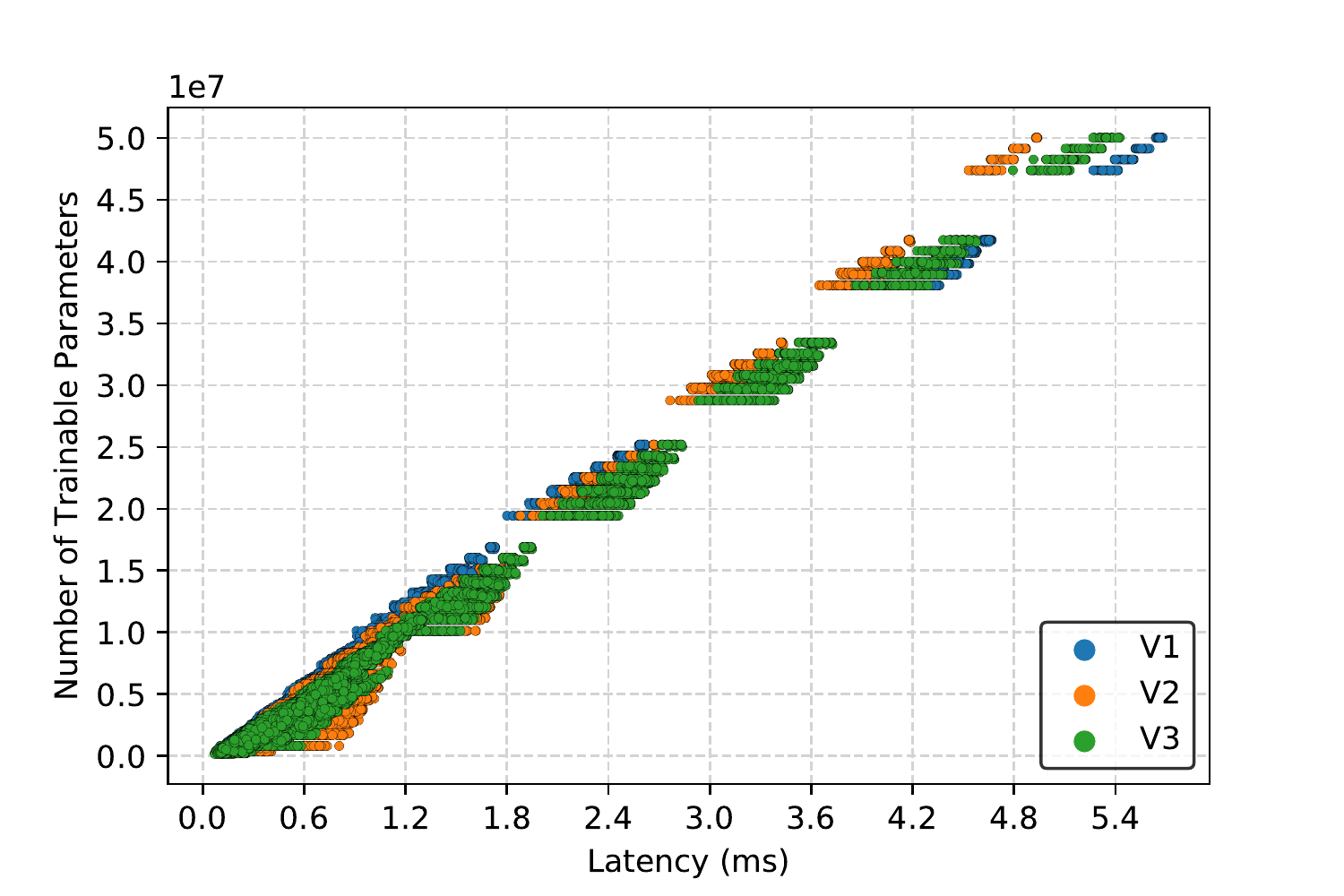}
    \caption{Scatter plot showing the number of trainable parameters in a NASBench model vs. the measured inference latency on three different configurations of the Google Edge TPU accelerators.}
    \label{fig:trainable_parameters_vs_latency}
\end{figure}
Figure~\ref{fig:trainable_parameters_vs_latency} shows the relationship between the number of trainable parameters in the NASBench models and their respective inference latency on three different configurations of the Edge TPU accelerator.
As increased number of trainable parameters comes with more MAC operations,
for all three configurations the latency is mostly directly proportional to the number of trainable parameters.
However, we observe that the ranking of the three configurations change at different model sizes. This is mainly explained by the interplay between parameter caching optimization (see Section~\ref{sec:sw}) and the available memory bandwidth. 

For very small models, all three configurations can cache the entire model which makes the overall latency very small.
For this region, overall the inference latencies are very close to each other for all configurations.
For the medium size models (5-30 million parameters), we observe that the \bench{V1} configuration provides the best performance. This is mainly attributed to its larger on-chip SRAM which can cache larger portions of the model whereas the other two configurations mostly end up streaming the parameters from off-chip. 

Interestingly, there is a cross-over point for larger models where the parameter caching has diminishing returns and parameters streamed from off-chip dominates the latency.
For these large models we observe that \bench{V2} and \bench{V3} provide better performance as they have larger memory bandwidth compared to \bench{V1}.
The difference between \bench{V2} and \bench{V3} configurations is mainly attributed to their architecture style.
Although their peak TOPS and bandwidth are the same, \bench{V2} achieves that with using more PEs whereas \bench{V3} uses less PEs but more cores per PE.
Having more PEs leads to fewer shared resources, less contention as well as more on-chip interconnect bandwidth, which enables \bench{V2} to sustain higher bandwidth.

\begin{figure*}[t]
    \centering
    \includegraphics[width=0.99\textwidth]{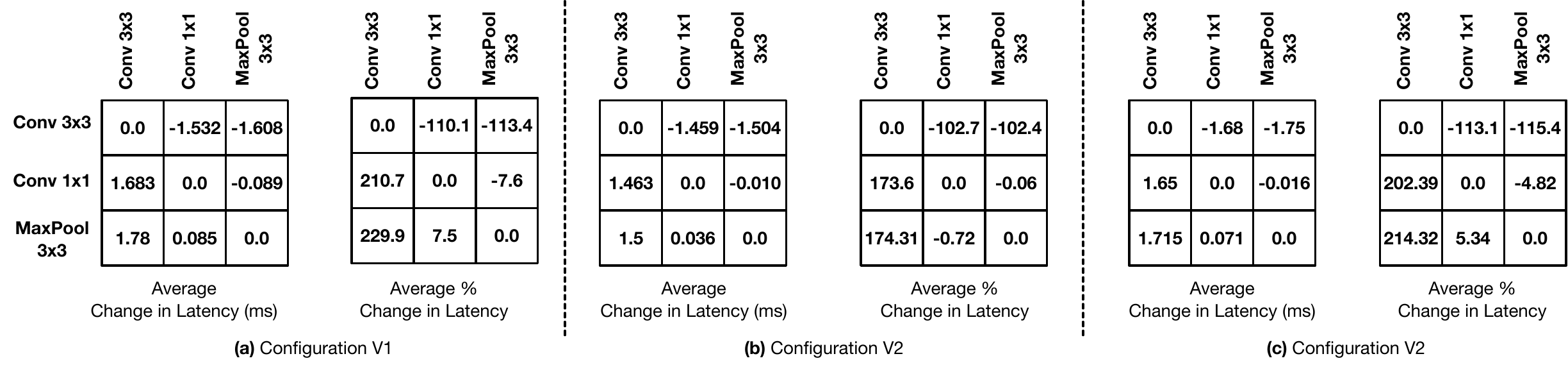}
    \caption{Measuring the aggregated impact of swapping cell operations on the performance of each class of accelerators. The rows show the original operation and the columns indicate the replacing operation in the NASBench cells. The first and second tables in each accelerator configuration present the average absolute and percentage change in latency, respectively.}
    \label{fig:op_impact}
\end{figure*}

\niparagraph{Performance and accuracy impact of operations.}
In Figure~\ref{fig:op_impact}, we investigate the effect of swapping each pair of the cell operations (3$\times$3 convolution, 1$\times$1 convolution, and 3$\times$3 max-pooling) on the performance of each class of accelerators.
For this swap, for every NASBench cell, we replace each cell operation with another operation to obtain a new set of cell operations.
Then, we search the NASBench dataset for a cell whose operations match the newly created cell operations and whose adjacency matrix matches the original cell's adjacency matrix.
The latency difference between these two cells are computed and averaged to obtain the latency impact of replacing cell operations. 
In a limited number of cases, the newly created cell operations does not match any existing cell in the the NASBench dataset, due to a mismatch between the number of input/output operations.
In such cases, the performance and mean validation accuracy difference measurement is skipped.

Figure~\ref{fig:op_impact} shows the latency change when operations are replaced with the aforementioned methodology.
Replacing a 1$\times$1 convolution with a 3$\times$3 increases the latency on all configurations but it increases the least (173.63\%) for the V2 architecture.
Changes in latency are not symmetric with respect to the swapping of cell operations.
For example, the latency reduction caused by changing a 1$\times$1 convolution to a 3$\times$3 convolution is not equal to the latency increase caused by changing a 3$\times$3 convolution to 1$\times$1 convolution.
This is because the changes in latency are measured by simulating and training the entire graph. 
The entire graph includes other operations and these other operations also contribute towards the graph's overall performance.

\begin{table}[t]
  \setlength{\tabcolsep}{2pt}
  \begin{center}
    \caption{Average accuracy, Spearman and Pearson correlations of the configuration specific learned performance estimation model}.
    \label{tab:summary:gcn_accuracy}
    \begin{tabular}{P{2.5cm}|P{1.5cm}|P{1.9cm}|P{1.9cm}|}
    \cline{2-4}
    &\textbf{V1}&\textbf{V2}&\textbf{V3}\\\hline
    \multicolumn{1}{|l|}{\textbf{Learning Rate}}&0.001&0.001&0.001\\\hline
    \multicolumn{1}{|l|}{\textbf{Batch Size}}&16&16&16\\\hline
    \multicolumn{1}{|l|}{\textbf{Training Set Size}}&254160&254160&254160\\\hline
    \multicolumn{1}{|l|}{\textbf{Test Set Size}}&84680&84680&84680\\\hline
    \multicolumn{1}{|l|}{\textbf{Avg. Accuracy}}&0.968&0.979&0.964\\\hline
    \multicolumn{1}{|l|}{\textbf{Spearman Correlation}}&0.99977&0.99981&0.99925\\\hline
    \multicolumn{1}{|l|}{\textbf{Pearson Correlation}}&0.99959&0.99974&0.99975\\\hline
    \end{tabular}
  \end{center}
\end{table}
\niparagraph{Accuracy of learned performance models.}
As mentioned in Section~\ref{sec:gnn}, we develop a learned model to estimate the various performance metrics of the accelerators.
In this section, we investigate the learned model accuracy in estimating the latency of NASBench model across three studied Edge TPU configurations. 
Table~\ref{tab:summary:gcn_accuracy} summarizes the training hyperparameters for the graph-based learned performance model.
In addition, it shows the performance of the learned model in predicting the inference latency of the NASBench models across the studied accelerator configurations. 
The average accuracy of the learned model in estimating the inference latency is around 96\%.
We also report both the Spearman rank-order correlation and Pearson linear correlation metrics of the learned performance model with the ground truth latency numbers for each accelerator configurations.
The results show that the learned model yields high correlation with the ground truth values ($>$~0.99).
This strong correlation, especially the Spearman rank-order correlation, signifies that the learned performance model is a strong candidate for replacing the expensive-to-evaluate cycle-accurate simulators in design space exploration and hardware/software co-optimization approaches.
This is because design space exploration and hardware/software co-optimization approaches only need to rank different configurations instead of measuring the absolute values.
\subsection{Architectural Insights for Edge TPUs}
\niparagraph{High-performing accelerator for large models.}
For models with larger graph depth and/or with more 3$\times$3 convolution operations (larger number of trainable parameters), accelerator configuration \bench{V2} yields lower latency as a result of its higher I/O bandwidth.
For these large models, in general, parameter caching does not help.
That is, when the number of trainable parameters are large, parameter caching leads to diminishing returns.
However, when the models have a smaller number of trainable parameters, parameter caching helps. This results in better performance by accelerator configuration V1.

\niparagraph{Impact of accelerator tile size on performance.}
For convolutional neural networks in the NASBench dataset, the latency is directly proportional to the number of trainable parameters.
Even though the inference latency of these neural models are dependent on the neural architecture or graph structure, the number of trainable parameters has higher impact on inference latency.
Therefore, it seems that I/O bandwidth is the deciding factor and other microarchitecture features (e.g. the number of PEs and compute cores) have lower impact on the overall accelerator latency.
To summarize, for the neural models in the NASBench dataset, we can reduce the accelerator tile size and still achieve a similar performance.
\section{Conclusion}
\label{sec:conclusion}
This paper evaluates three different classes of Edge TPU accelerators, covering various computing ecosystems, across more than 423K unique convolutional neural networks.
Analyzing these results, we draw critical and interpretable microarchitectural insights that help understand the architectural trade-offs in Edge TPUs.
Finally, we discuss our proposed robust learned models to estimate the major performance metrics of Edge TPU accelerators.
We show that the graph-based learned performance model estimates the latency and energy consumption of three studied classes of Edge TPUs with around 97\% accuracy and high correlations ($>$99$\%$) with the ground truth data.
This high-accuracy learned model paves the way for rapid architecture exploration and hardware/software co-design, which we leave as future work.
\section*{Acknowledgment}
We would like to extend our gratitude towards Suyog Gupta, Samy Bengio, Cliff Young, Chandu Thekkath, Stella Aslibekyan, the ``Learn to Design Accelerators'' team, and the extended Google Research Brain Team for their invaluable feedback and comments.
\section*{}
\bibliographystyle{IEEEtranS}
\bibliography{paper}

\end{document}